# Seabed-Net: A multi-task network for joint bathymetry estimation and seabed classification from remote sensing imagery in shallow waters

Panagiotis Agrafiotis [a,b,*], Begüm Demir [a,b]

[a]Faculty of Electrical Engineering and Computer Science, Technische Universität Berlin, Berlin, 10587, Germany
[b]BIFOLD - Berlin Institute for the Foundations of Learning and Data, Berlin, 10587, Germany

ABSTRACT

Accurate, detailed, and regularly updated bathymetry, coupled with complex semantic content, is essential for under-mapped shallow-water environments facing increasing climatological and anthropogenic pressures. However, existing approaches that derive either depth or seabed classes from remote sensing imagery treat these tasks in isolation, forfeiting the mutual benefits of their interaction and hindering the broader adoption of deep learning methods. To address these limitations, we introduce Seabed-Net, a unified multi-task framework that simultaneously predicts bathymetry and pixel-based seabed classification from remote sensing imagery of various resolutions. Seabed-Net employs dual-branch encoders for bathymetry estimation and pixel-based seabed classification, integrates cross-task features via an Attention Feature Fusion module and a windowed Swin-Transformer fusion block, and balances objectives through dynamic task uncertainty weighting. In extensive evaluations at two heterogeneous coastal sites, it consistently outperforms traditional empirical models and traditional machine learning regression methods, achieving up to 75% lower RMSE. It also reduces bathymetric RMSE by 10-30% compared to state-of-the-art single-task and multi-task baselines and improves seabed classification accuracy up to 8%. Qualitative analyses further demonstrate enhanced spatial consistency, sharper habitat boundaries, and corrected depth biases in low-contrast regions. These results confirm that jointly modeling depth with both substrate and seabed habitats yields synergistic gains, offering a robust, open solution for integrated shallow-water mapping. Code and pretrained weights are available at https://github.com/pagraf/Seabed-Net.

## 1. Introduction

Water covers approximately 71% of the Earth's surface, with oceans accounting for about 96.5% of this area - roughly 362 million square kilometers. Yet, only a small portion has been directly mapped in high resolution. The growing need for detailed knowledge of underwater terrain, including seabeds, riverbeds, and lakebeds, is driven by factors such as habitat destruction, marine pollution, threats to submerged cultural heritage, recent maritime tragedies, natural disasters, navigation safety, and the rising demand for offshore energy and marine resources. For shallow coastal areas, especially impacted by environmental and human pressures, accurate and detailed bathymetric data, along with the respective seabed class information (i.e., semantic information about benthic habitats and substrates such as sand, seagrass, or rock), are crucial.

Conventional seabed mapping techniques, such as echosounders, are often inefficient in shallow water environments, where they are hindered by wave-induced interference, the presence of reefs, and multi-path propagation errors. Although airborne LiDAR (Light Detection and Ranging) offers an alternative for shallow water surveying, it remains costly and generally provides limited information about the seabed classes (Agrafiotis et al., 2020). To overcome these limitations, the use of aerial and satellite remote sensing imagery has gained increasing popularity for shallow water mapping (Mandlburger, 2022; Wang and Li, 2023), supporting both bathymetry retrieval and pixel-based seabed classification.

### 1.1. Bathymetry retrieval

For bathymetry retrieval, Spectrally-Derived Bathymetry (SDB) methods provide a cost-effective alternative to traditional acoustic and LiDAR surveys, especially in shallow coastal waters. These techniques leverage multispectral satellite imagery to estimate water depth and are increasingly used due to their wide spatial coverage and accessibility. The choice of the SDB approach depends on several factors, including the spatial extent of the study area, the desired resolution and accuracy, budget constraints, the availability of ground-truth data and the intended application of the bathymetric information.

SDB methods estimate water depth by modeling the attenuation of radiance as a function of depth and wavelength, based on the principle that light is progressively absorbed and scattered as it penetrates the water column. These methods exploit spectral variations captured in multispectral imagery to infer depth, particularly in optically shallow regions. Depending on the algorithm used, areas with relatively homogeneous seabed types often yield more accurate results, making SDB particularly suitable for large-scale mapping where high-resolution detail is less critical.

Traditional empirical (Stumpf et al., 2003; Lyzenga, 1978; Lyzenga et al., 2006), physics-based (Dekker et al., 2011; Hedley et al., 2012), and traditional machine learning regression-based methods such as random forests and support vector regression (Traganos et al., 2018; Sagawa et al., 2019; Misra and Ramakrishnan, 2020; Eugenio et al., 2021;

*Corresponding author
agrafiotis@tu-berlin.de (P.A. ); demir@tu-berlin.de (B.D. )
ORCID(s): 0000-0003-4474-5007 (P.A. ); 0000-0003-2175-7072 (B.D. )

Thomas et al., 2022; Mudiyanselage et al., 2022; Xie et al., 2023; Hao Quang et al., 2025), have been widely applied in SDB tasks. Empirical methods are grounded in radiative transfer theory, leveraging the exponential decay of light intensity with depth and the varying attenuation characteristics across different wavelengths. Their simplicity is a key advantage: they are typically defined by a single equation with only a few tunable parameters, making them easy to implement (Knudby and Richardson, 2023). Traditional regression methods build a single predictive model that estimates water depth directly from the spectral information of individual pixels, with parameters optimized during a calibration phase. These models tend to outperform traditional empirical approaches (Zhang et al., 2022b; Abdul Gafoor et al., 2022; Duan et al., 2022; Zhou et al., 2023b; Xie et al., 2023; Fang et al., 2024; Hao Quang et al., 2025; Cheng et al., 2025), as they are more flexible to capture more complex relationships between spectral inputs and depth. A limited number of studies have also explored spatially adaptive techniques, such as geographically weighted regression (Liu et al., 2018; Fang et al., 2024) which allow model parameters to vary smoothly across geographic areas to better account for local environmental differences.

These approaches are now increasingly outperformed by deep learning methods (Ceyhun and Yalçın, 2010; Ai et al., 2020; Lumban-Gaol et al., 2021; Al Najar et al., 2021; Kaloop et al., 2021; Lumban-Gaol et al., 2022; Peng et al., 2022; Mandlburger et al., 2021; Xi et al., 2023; Knudby and Richardson, 2023; Shen et al., 2023; Zhou et al., 2023a; Gupta et al., 2024; Chen et al., 2024; Wu et al., 2024; Qin et al., 2024; Agrafiotis and Demir, 2025; Cheng et al., 2025; Lv et al., 2025; García-Díaz et al., 2025; Lee et al., 2025; Zhu et al., 2025), which have gained significant traction in SDB for their ability to model the complex interactions of light with the water surface, the water column, and the seabed. This shift reflects the superior capacity of deep networks to capture nonlinear spectral-spatial patterns, particularly in optically complex environments with heterogeneous seabed compositions (Mandlburger, 2022).

However, these single-task bathymetry estimation models also face significant limitations when applied in diverse coastal environments. These models rely on image radiometry, which is highly sensitive to variations in water clarity, substrate reflectance, and sensor resolution. As a result, their performance tends to degrade sharply when applied to coarser-resolution data (e.g., Sentinel-2), or when generalizing across different geographic domains. Moreover, single-task bathymetry networks lack complementary semantic supervision that could help reinforce spatial consistency and suppress noise in ambiguous or low-contrast regions. Without auxiliary tasks such as seabed classification, these models often fail to capture meaningful context or structure, resulting in over-smoothed or spatially incoherent predictions. In addition, their exclusive focus on a single regression objective limits their ability to learn disentangled representations that generalize well across remote sensing sensors. Overall, the absence of multi-task contextualization makes single-task bathymetry models brittle, resolution-dependent, and less robust to complex marine conditions.

## 1.2. Seabed classification

Seabed habitats show significant variability in marine environments, driven by diverse factors such as water temperature, light availability, and seawater acidity. Traditional machine learning approaches, which rely on empirically designed features such as color, texture, and shape descriptors, often struggle to achieve robust performance under these complex and dynamic conditions (Chen et al., 2024). As a result, producing accurate and scalable seabed maps remains a considerable challenge. Although deep learning-based methods have demonstrated improved capability in pixel-based classification of the seabed (Huang et al., 2022; B. Lyons et al., 2020; Chen et al., 2023; Agrafiotis et al., 2024) by learning from sparse manually annotated data or in situ measurements, they still face several critical limitations.

In particular, single-task classification networks primarily focus on learning spectral or appearance-based features, without explicit modeling of the structural or geometric organization of benthic habitats. This shortcoming is magnified under sparse supervision, where the absence of detailed structural labels prevents the recovery of fine-scale habitat boundaries and topological relationships. Moreover, current single-task methods typically do not exploit auxiliary data, such as bathymetric information, to improve structural reasoning. As a consequence, pixel-based classification output often suffers from poor spatial consistency, oversmoothing, and susceptibility to environmental artifacts such as water column variability and sun glint (Chen et al., 2024, 2025). Addressing these challenges requires multi-task frameworks capable of integrating semantic, geometric, and spectral features to improve classification robustness and structural accuracy across diverse sensing conditions.

## 1.3. Combined approaches

Given these challenges, bathymetry and pixel-based classification can support each other; On the one hand, bathymetry can provide important spatial and geometric features which are invaluable for seabed classification as they capture abrupt changes in the seabed structure, which often coincide with shifts in seabed classes. The spectral differences of benthic habitats are visible in remote sensing images because of bathymetry, with bathymetric features providing additional contextual and local/ global geometric data that enrich model classification. One the other hand, seabed classification can significantly improve bathymetric accuracy by helping to distinguish between different seabed types that may affect depth perception in imagery; certain seabed classes, like seagrass or algae-covered areas, often appear darker in satellite or aerial images, which can lead algorithms to misinterpret these areas as deeper than they actually are. Incorporating seabed classification can help bathymetric models adjust for this by recognizing specific spectral patterns associated with various seabed classes, such as differentiating between truly deep areas and areas covered by vegetation.

Multi-task architectures (Xu et al., 2018; Vandenhende et al., 2020; Ekim and Sertel, 2021; Zhang et al., 2022a; Ye and Xu, 2024) enable tasks like semantic segmentation, instance segmentation, edge detection, and depth estimation to work together, enhancing the model's understanding of image content. By learning multiple tasks within one model, information from different tasks can be shared complementing each other. This strategy often leads to a more effective model than single-task training (Ye and Xu, 2024) and suggests that jointly learning bathymetry and seabed classification can enable richer representations and improved seabed mapping outcomes.

While such approaches have gained traction in terrestrial remote sensing (Carvalho et al., 2019; Wang et al., 2020; Zhao et al., 2023), their use for shallow-water seabed mapping remains largely unexplored, failing to address the unique challenges of the underwater domain. Some early efforts have attempted to bridge this gap through bathymetry-guided classification (Gupta et al., 2024; Chen et al., 2024, 2025) and class-aware inversion strategies (Zuo et al., 2025). They typically rely on handcrafted fusion schemes that may fail to preserve the integrity of task-specific characteristics or generalize across sensing conditions. They focus primarily on improving classification accuracy, treating bathymetry as a secondary objective. In addition, they fail to address the complex trade-offs involved in joint training, such as cross-task interference and feature entanglement between semantic and geometric representations. As such, a fully integrated framework that simultaneously and jointly optimizes both bathymetry estimation and seabed classification, while maintaining robust task disentanglement, remains largely unexplored. This gap highlights the pressing need for architectures capable of deep feature integration and balanced multi-task optimization to improve seabed mapping in complex shallow water environments.

### 1.4. Contributions

To tackle these challenges across varying spatial resolutions and sensors, we introduce a novel multi-task learning framework that jointly estimates continuous bathymetry and performs pixel-based seabed classification. Our model is designed to leverage shared spatial and spectral representations between tasks, improving generalization and robustness, especially under lower-resolution satellite data where conventional single-task models tend to degrade significantly. The main contributions of this work are summarized as follows:

- We propose Seabed-Net, the first dual-branch multi-task architecture for jointly performing bathymetry retrieval and pixel-level seabed classification. By learning shared representations, the two tasks mutually enhance each other: depth cues refine class boundaries, while habitat information improves depth estimation.

- To jointly optimize bathymetry and pixel-based classification tasks, we adopt a multi-task learning strategy based on task uncertainty weighting, as proposed in Kendall et al. (2018).

- We conduct a comprehensive evaluation using imagery from three different sensors of various spatial resolutions over diverse coastal settings.

- We demonstrate that our method outperforms state-of-the-art single-task and multi-task baselines in both bathymetry and seabed classification accuracy, particularly under challenging low-resolution settings.

- We provide detailed ablation and cross-modal performance analyses, highlighting the resilience of our multi-task framework to resolution degradation and its effectiveness in learning shared representations across tasks and domains.

## 2. Proposed method

The proposed Seabed-Net architecture (Fig. 1) employs two parallel encoders - one for each task, built from convolutional blocks, and two corresponding decoders tailored for bathymetry and pixel-based classification outputs. To facilitate task interaction and cross-representation learning, we introduce two complementary fusion mechanisms: (i) an Attention Feature Fusion (AFF) module (Fig. 2) that combines spatial and channel-wise dependencies between task-specific features to emphasize locally relevant features, and (ii) a Vision Transformer (ViT) (Dosovitskiy, 2020) Fusion module (Fig. 3) that leverages patch embeddings, cross attention, and local self-attention within windowed regions to capture global context. At each encoder level, the corresponding feature maps from the two branches are fused using the two specialized modules, enabling cross-task information exchange at multiple scales, enriching each task's feature representation with complementary information from the other. These components allow Seabed-Net to selectively share and refine information across tasks and spatial scales, improving both accuracy and robustness across sensors.

### 2.1. Task-specific encoders

Given an input image $X \in \mathbb{R}^{B \times 3 \times H \times W}$, where $B$ is the batch size, two separate encoders are applied: one for bathymetry and one for pixel-based classification. Each encoder produces four feature maps:

$$\{e_b^{(i)}\}_{i=1}^4 = \mathcal{E}_b(X), \tag{1}$$

$$\{e_s^{(i)}\}_{i=1}^4 = \mathcal{E}_s(X), \tag{2}$$

where $\mathcal{E}_b(\cdot)$ and $\mathcal{E}_s(\cdot)$ denote the bathymetry and pixel-based classification encoders, respectively, and $e_b^{(i)}$, $e_s^{(i)}$ represent the feature maps at the $i$-th scale.

Four convolutional blocks are defined to progressively extract features from the input images. Each block utilizes a series of two convolutional layers followed by ReLU activation functions. In each block, the number of filters doubles with each successive layer: from 64 to 128, 128 to 256, and 256 to 512. The bathymetry encoder omits batch normalization (BN) layers to retain fine radiometric information, important for bathymetry prediction, whereas the pixel-based classification encoder includes batch normalization to enhance feature generalization for classification.

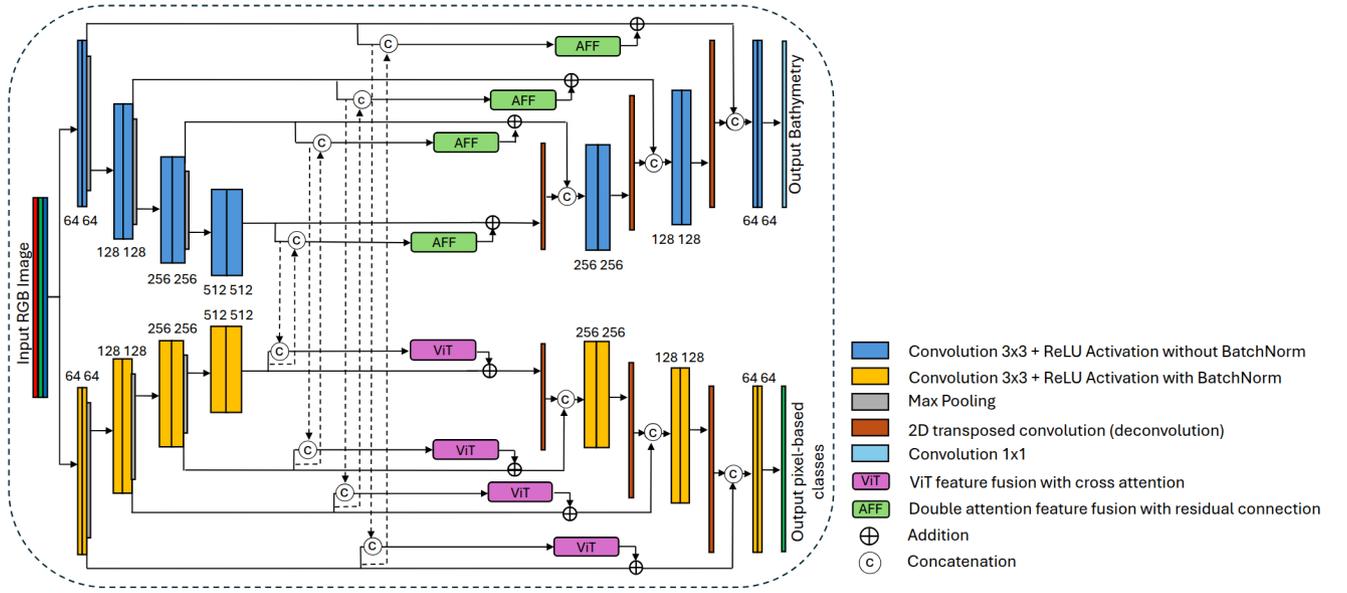

**Figure 1:** The proposed Seabed-Net architecture.

## 2.2. Attention feature fusion (AFF)

At each encoder scale $i$, local cross-task interactions are captured using Attention Feature Fusion (AFF) modules (Fig. 2). The bathymetry and pixel-based classification features are first concatenated and passed through a spatial attention mechanism $\mathcal{A}_{\text{spatial}}$ and a channel attention mechanism $\mathcal{A}_{\text{channel}}$.

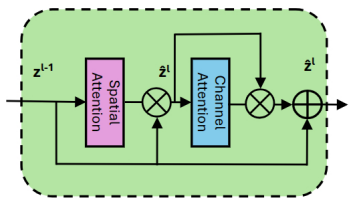

**Figure 2:** The double attention feature fusion (AFF) block.

The fused feature at scale $i$ is given by:

$$\begin{aligned} f_{\text{AFF}}^{(i)} &= \mathcal{F}_{\text{AFF}}\left(e_b^{(i)}, e_s^{(i)}\right) \\ &= \text{Conv1x1}\left((e_b^{(i)} \oplus e_s^{(i)}) \odot \hat{z}^l \odot z^l\right) \\ &\quad + (e_b^{(i)} \oplus e_s^{(i)}), \end{aligned} \quad (3)$$

where $\oplus$ denotes channel-wise concatenation and $\otimes$ indicates element-wise multiplication. The spatial attention map $\hat{z}^l$ is defined by:

$$\hat{z}^l = \sigma\left(\text{Conv7x7}\left(e_b^{(i)} \oplus e_s^{(i)}\right)\right), \quad (4)$$

where $\sigma(\cdot)$ is the sigmoid function and Conv7x7 represents a $7 \times 7$ convolution that captures spatial dependencies across the concatenated feature map.

The channel attention map $z^l$ is computed using a squeeze-and-excitation (SE) block. First, global average pooling is applied to the input feature map to produce a channel descriptor. Then, two successive $1 \times 1$ convolutions with a ReLU activation in between are used to generate channel-wise weights:

$$z^l = \sigma\left(W_2, \delta\left(W_1 \cdot \text{GAP}(e_b^{(i)} \oplus e_s^{(i)})\right)\right), \quad (5)$$

where $\text{GAP}(\cdot)$ denotes global average pooling, $W_1$ and $W_2$ represent the weights of the first and second $1 \times 1$ convolution layers respectively, $\delta(\cdot)$ is the ReLU activation, and $\sigma(\cdot)$ is the sigmoid function.

A residual connection is added to preserve the original feature information before applying a $1 \times 1$ convolution to match the output dimensionality. The resulting fused feature $f_{\text{AFF}}^{(i)}$ is then used to guide the bathymetry decoding path by emphasizing localized and semantically relevant regions important for depth estimation.

## 2.3. ViT-based feature fusion

In parallel, the global context is modeled through a Vision Transformer (ViT) fusion module at each scale (Fig. 3). Here, the bathymetry and pixel-based classification features are concatenated and projected into an embedding space, processed with a lightweight Swin Transformer (Liu et al., 2021) block $\mathcal{T}$, and then reprojected to a lower dimension:

$$f_{\text{ViT}}^{(i)} = \mathcal{F}_{\text{ViT}}\left(e_b^{(i)}, e_s^{(i)}\right)$$
$$= \text{Conv1x1}_{\text{out}}\left(\mathcal{T}\left(\text{Conv1x1}_{\text{in}}\left(e_b^{(i)} \oplus e_s^{(i)}\right)\right)\right), \quad (6)$$

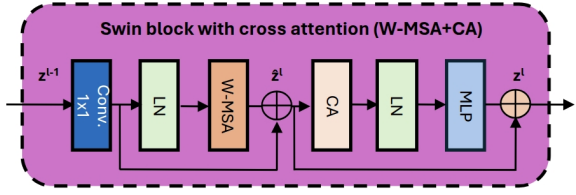

**Figure 3:** The ViT block.

The Swin Transformer block is composed of a window-based multi-head self-attention layer, a cross-attention layer, and a multilayer perceptron (MLP) with ReLU activation. Initially, the input is normalized using Layer Normalization (LN). The block applies attention within non-overlapping local windows, enabling localized self-attention operations. The cross-attention (CA) layer facilitates information exchange across different input sequences. The output is then normalized again and passed through the MLP, with residual connections used to integrate the intermediate results. The Swin Transformer blocks are computed as:

$$\hat{z}^l = \text{W-MSA}(\text{LN}(z^{l-1})) + z^{l-1},$$
$$z^l = \text{MLP}(\text{LN}(\text{CA}(\hat{z}^l))) + \hat{z}^l \quad (7)$$

There, $\hat{z}^l$ and $z^l$ denote the output features of the W-MSA module and the CA and MLP module for block $l$, respectively; W-MSA denotes window based multi-head self-attention using regular window partitioning configuration.

The W-MSA mechanism computes the relationships between positions in the input by projecting it into query, key, and value matrices:

$$Q = XW_Q, \quad K = XW_K, \quad V = XW_V, \quad (8)$$

where $W_Q, W_K, W_V \in \mathbb{R}^{D \times D}$ are the learned weights. The attention output is then computed as:

$$\text{W-MSA}(Q, K, V) = \text{Softmax}\left(\frac{QK^\top}{\sqrt{d_k}}\right)V, \quad (9)$$

with $d_k = \frac{D}{N_{\text{heads}}}$ being the dimensionality of the output of each head, serving as a scaling factor to stabilize the gradients during training.

To capture interactions between features at different levels, a cross-attention (CA) mechanism is also employed. Given high-level features $X_{\text{high}} \in \mathbb{R}^{B \times N_{\text{high}} \times D}$ and low-level features $X_{\text{low}} \in \mathbb{R}^{B \times N_{\text{low}} \times D}$, the projections are defined as:

$$Q = X_{\text{high}} W_Q, \quad K = X_{\text{low}} W_K, \quad V = X_{\text{low}} W_V, \quad (10)$$

and the resulting cross-attention output is computed by:

$$\text{CA}(Q, K, V) = \text{Softmax}\left(\frac{QK^\top}{\sqrt{d_k}}\right)V. \quad (11)$$

The MLP is defined as:

$$\text{MLP}(X) = W_2 \cdot \text{ReLU}(W_1 \cdot X + b_1) + b_2 \quad (12)$$

where, $W_1 \in \mathbb{R}^{D \times D_{\text{hidden}}}$ is the weight matrix for the first linear transformation, $W_2 \in \mathbb{R}^{D_{\text{hidden}} \times D}$ is the weight matrix for the second linear transformation, $b_1 \in \mathbb{R}^{D_{\text{hidden}}}$ and $b_2 \in \mathbb{R}^D$ are the bias terms, $D$ is the input feature dimension, $D_{\text{hidden}} = 4D$ is the dimensionality of the hidden layer. Here, $X$ is the input tensor, and after the first linear transformation $W_1 \cdot X + b_1$, the output is passed through the ReLU activation function. The resulting tensor is then transformed by $W_2$ to produce the final output of the MLP.

The ViT fusion outputs are designed to support the pixel-based classification decoder by providing enhanced global spatial coherence between regions.

### 2.4. Task-specific decoders

Each encoder level produces two separate fused outputs: the AFF-derived feature map and the ViT-derived feature map. These are routed into the decoders in a task-specific manner: the AFF output is used as a skip connection in the bathymetry decoder, whereas the ViT output is used in the pixel-based classification decoder. In practice, before addition, the fused features are bilinearly interpolated to match the spatial dimensions of the corresponding encoder outputs.

The bathymetry decoder $\mathcal{D}_b$ reconstructs the continuous depth map $\hat{Y}_b$ using the AFF-fused features:

$$\hat{Y}_b = \mathcal{D}_b\Big(f_{\text{AFF}}^{(1)} + e_b^{(1)}, f_{\text{AFF}}^{(2)} + e_b^{(2)},$$
$$f_{\text{AFF}}^{(3)} + e_b^{(3)}, f_{\text{AFF}}^{(4)} + e_b^{(4)}\Big). \quad (13)$$

For pixel-based classification decoding, the ViT-fused features and pixel-based classification encoder features are combined:

$$\hat{Y}_s = \mathcal{D}_s\Big(f_{\text{ViT}}^{(1)} + e_s^{(1)}, f_{\text{ViT}}^{(2)} + e_s^{(2)},$$
$$f_{\text{ViT}}^{(3)} + e_s^{(3)}, f_{\text{ViT}}^{(4)} + e_s^{(4)}\Big). \quad (14)$$

Both decoders are composed of successive upsampling (transposed convolution) blocks and skip connections from the encoder. The pixel-based classification decoder additionally applies batch normalization after each convolution to stabilize the multi-class classification task, whereas the bathymetry decoder omits normalization layers to preserve the precision required for continuous depth regression.

In summary, the Seabed-Net architecture leverages both localized (via AFF) and global (via ViT) cross-task feature interactions while preserving task-specific pathways during decoding. This design promotes efficient learning of shared representations while enabling specialization for each task's prediction objective.

### 2.5. Objective function

To jointly optimize the bathymetry and pixel-based classification tasks, we adopt a multi-task learning strategy based on task uncertainty weighting, as proposed in Kendall et al. (2018). Instead of manually setting fixed loss weights, we learn task-specific uncertainties during training to dynamically balance the contributions of each task. This approach prevents dominance of one task over the other, ensuring stable and effective joint learning.

Specifically, the overall loss $\mathcal{L}_{\text{total}}$ is formulated as:

$$\mathcal{L}_{\text{total}} = \frac{1}{2\sigma_b^2} \mathcal{L}_{\text{bathymetry}} + \frac{1}{2\sigma_s^2} \mathcal{L}_{\text{classification}} \quad (15)$$
$$+ \log \sigma_b + \log \sigma_s,$$

where $\mathcal{L}_{\text{bathymetry}}$ is the task-specific loss for depth prediction, $\mathcal{L}_{\text{classification}}$ is the task-specific loss for seabed classification, and $\sigma_b$, $\sigma_s$ are learnable scalar parameters representing the task uncertainties.

The bathymetry task is supervised using the Root Mean Squared Error (RMSE) loss:

$$\mathcal{L}_{\text{bathymetry}} = \sqrt{\frac{1}{N} \sum_{i=1}^{N} (Y_b(i) - \hat{Y}_b(i))^2}, \quad (16)$$

while the classification task is supervised using the standard cross-entropy loss:

$$\mathcal{L}_{\text{classification}} = -\frac{1}{N} \sum_{i=1}^{N} \sum_{c=1}^{C} \hat{Y}_s(i, c) \log Y_s(i, c), \quad (17)$$

where $N$ denotes the number of pixels and $C$ the number of classes.

The introduction of the learnable uncertainty parameters allows the model to automatically attenuate the influence of each task's loss based on its inherent difficulty, resulting in a more stable and balanced optimization.

## 3. Dataset and experimental setup

### 3.1. Dataset description

To evaluate the proposed model using imagery from different sensors and different areas, we used the MagicBathyNet dataset (Agrafiotis et al., 2024). MagicBathyNet's coverage includes two vastly different coastal areas in terms of water column characteristics and bottom type: (i) Agia Napa area in Cyprus (Fig. 4a), which represents a diverse range of typical Mediterranean coastal waters and seabed compositions, including rock, sand, seagrass (Posidonia oceanica), and macroalgae (Filamentous/turf algae); and (ii) Puck Lagoon area in Poland (Fig. 4b), which is a characteristic lagoon of the Baltic Sea, predominantly featuring eelgrass/pondweed (Zostera marina, Stuckenia pectinata, Potamogeton perfoliatus) and sand, while also encompassing ports, marinas, estuaries, and dredged areas.

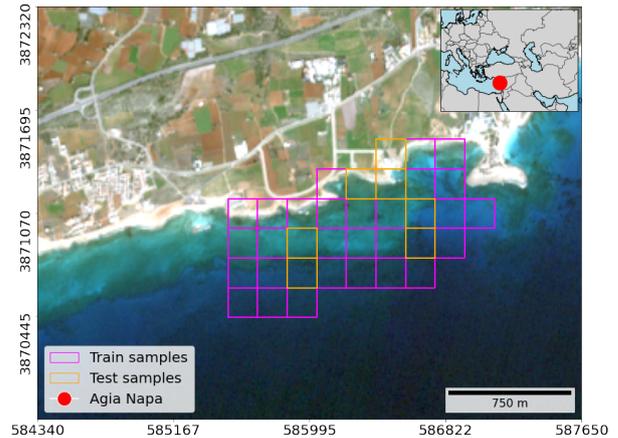

(a)

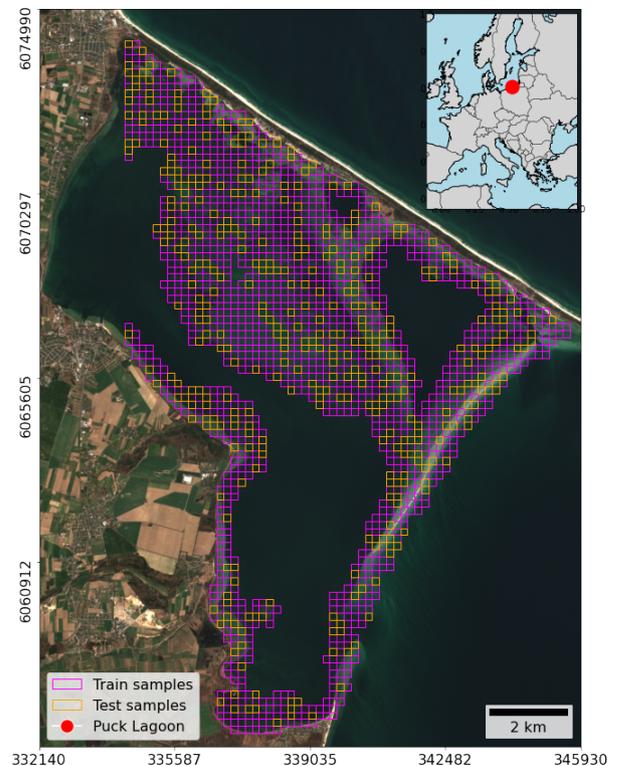

(b)

**Figure 4:** The considered areas: (a) Agia Napa and (b) Puck Lagoon as depicted in the Sentinel-2 Level2A imagery (source: Copernicus Hub). For Agia Napa, the coordinate reference system EPSG:32636 - WGS 84 / UTM zone 36N is used, while for Puck Lagoon, EPSG:32634 - WGS 84 / UTM zone 34N is applied.

MagicBathyNet contains co-registered triplets of Sentinel-2 (S2) Level2A, SPOT 6 (ORTHO prodcut), and aerial orthoimage patches, complemented by Digital Surface Models (DSM) of the seabed and annotated patches for seabed habitat and type, facilitating bathymetry prediction and supervised pixel-based classification respectively. All the considered satellite images were radiometrically, geometrically, and atmospherically corrected. No sun glint removal was performed on any of the images, as our models were aimed at operating directly on real-application image data in a single end-to-end processing step. We used only RGB bands for each sensor across all models to ensure consistency between sensors and maintain generalization in real-world deployments. This choice was made intentionally to reflect the most commonly available spectral configuration in operational coastal mapping workflows, especially for aerial imagery where additional bands may not always be accessible. In Agia Napa, the seabed of the area used in this work reaches a depth of -14 m. Bathymetric LiDAR data were acquired using the Leica HawkEye III system at a density of 1.5 points/$m^2$. The aerial image collection was conducted using a fixed-wing UAV equipped with a Canon IXUS 220HS camera, featuring a 4.30mm focal length, 1.55$\mu$m pixel size, and a 4000×3000 pixels format. The selected area of the Puck Lagoon reaches a depth of -8 m. Bathymetric data obtained using the Riegl VQ-880-GII LiDAR and Teledyne Reson T50/T20 multibeam echosounders at a density of 12 to 25 points/$m^2$. Aerial imagery was obtained using a Phase One iXM-100 camera with a 35mm focal length, 3.7$\mu$m pixel size, and a 11664x8750 pixels format. Details about the SfM-MVS steps using the aerial data, number of GCPs used, remaining errors in the bundle adjustment, DSM generation process as well as data acquisition dates are provided in (Agrafiotis et al., 2024).

For each area and remote sensing sensor, approximately 700 pixels (samples) were manually annotated by experts, informed by geolocated seabed samples collected through underwater imagery and probe measurements. This annotation process was carried out as part of the MagicBathyNet (Agrafiotis et al., 2024) creation effort (Table 1). The annotations covered five different classes: i) seagrass (Posidonia oceanica); ii) macroalgae (Filamentous/turf algae); iii) eelgrass/pondweed (Zostera marina, Stuckenia pectinata, Potamogeton perfoliatus); iv) sand; and v) rock. It is noted that the Puck Lagoon area primarily contains classes that are also covered by algal mats, resulting in the prevalent greenish hues seen in most of the annotated pixels. Each patch covers an area of 180x180m, represented by 18x18 pixels in Sentinel-2 imagery (10 m spatial resolution), 30x30 pixels in SPOT 6 imagery (6 m spatial resolution) and 720x720 pixels in airborne imagery (0.25 m spatial resolution).

### 3.2. Baselines and compared methods

To assess the effectiveness of the proposed SeabedNet architecture, we conduct a comprehensive comparison against empirical (Emp.), traditional machine learning regression methods (Regr.), single-task and multi-task deep learning baselines using the aerial, SPOT 6 and Sentinel-2 imagery.

**Table 1**
The number of samples per class in MagicBathyNet (Agrafiotis et al., 2024).

| Class | Agia Napa | Puck Lagoon | Total samples |
|---|---|---|---|
| Seagrass | 118 | 0 | 118 |
| Macroalgae | 82 | 0 | 82 |
| Eelgrass etc. | 0 | 428 | 428 |
| Sand | 246 | 326 | 572 |
| Rock | 251 | 0 | 251 |
| # of Samples | 697 | 754 | 1451 |
| # of Patches | 35 | 498 | - |

The empirical models tested in this study-Multiband Linear (Lyzenga, 1978) and Log Band Ratio (Stumpf et al., 2003)-represent widely used, physics-inspired methods for SDB. These models use simple regression-based relationships between spectral bands to estimate water depth and have long served as standard baselines in the field. Specifically, the Log Band Ratio model uses only the blue and green bands, while the Multiband Linear model employs all three RGB bands. In addition, traditional data-driven machine learning models such as Random Forest Regression (with 300 trees) and Support Vector Regression (SVR) with a radial basis function (RBF) kernel were also evaluated, both using all three RGB bands as input features. These models capture non-linear relationships between spectral features and depth and have been increasingly adopted as robust baselines for SDB estimation in recent years.

Among single-task deep learning-based approaches, a UNet-bathy variant (Mandlburger et al., 2021; Agrafiotis et al., 2024) and UNet (Ronneberger et al., 2015) serve as baselines for bathymetry estimation and pixel-based classification, respectively, while SegFormer (Xie et al., 2021) is evaluated solely on pixel-based classification performance. These models are trained to perform semantic pixel-based classification of different seabed types and are not designed to predict continuous bathymetry values. Their inclusion in the comparison serves to assess the potential of traditional pixel-based classification architectures when used exclusively for classification tasks within the same coastal environments.

For multi-task learning, we compare against several state-of-the-art deep learning-based models, including PAD-Net (Xu et al., 2018), MTI-Net (Vandenhende et al., 2020), MTL (Ekim and Sertel, 2021), JSH-Net (Zhang et al., 2022a), and TaskPrompter (Ye and Xu, 2024). These models represent strong baselines that aim to exploit cross-task synergies to improve performance across sensors.

### 3.3. Experimental setup

We implemented our method in PyTorch using one NVIDIA A100 80GB GPU. For training, we employed the combined loss function (Eq. 15) alongside the Adam optimizer (Kingma, 2014), with a cosine annealing learning rate scheduler set to decay the learning rate over a cycle of 10 epochs. Each epoch consisted of 10.000 iterations, resulting in a total of 100.000 gradient updates. The initial learning rate was $10^{-4}$ for SPOT 6 and Sentinel-2, and $10^{-5}$ for aerial data for a 10-epoch training period. Data normalized to [0, 1] range by dividing the pixels values of the RGB orthoimages with 255 and the depths with -14.556 m for the Agia Napa area and with -11 m for the Puck Lagoon area. For training the models with Sentinel-2 and SPOT 6 data, for both tasks, resized 256x256 crops were used, notably improving performance without adding new information. As for the data augmentation, we performed random rotations, as well as random vertical and horizontal flips.

To ensure reproducibility, for each remote sensing sensor and geographic area in the dataset we used the predefined 80%-20% splits available in MagicBathyNet dataset (Agrafiotis et al., 2024). In the Agia Napa region, this resulted in a total of 35 annotated patches, with 28 used for training and 7 for evaluation. In the larger Puck Lagoon area, 1971 annotated patches were available, of which 1575 were allocated for training and 396 for evaluation. All test patches were strictly withheld during training, ensuring that the model was evaluated only on unseen data.

### 3.4. Metrics used

To assess the models' performance quantitatively, we relied on five evaluation metrics: i) Root Mean Squared Error (RMSE), which captures the average magnitude of depth prediction error with higher sensitivity to large errors (Eq. 18); ii) Mean Absolute Error (MAE), which measures the average absolute difference between predicted and reference depths (Eq. 19); iii) Standard Deviation (Std), indicating the variability of the depth prediction error (Eq. 20); iv) Overall Accuracy (OA), which measures the proportion of correctly classified pixels (Eq. 21); and v) mean Intersection over Union (mIoU), which reflects the average overlap between predicted and ground truth classes across all classes, and is particularly sensitive to class-level performance and imbalance (Eq. 22). The evaluation metrics were calculated as follows:

$$\text{RMSE} = \sqrt{\frac{1}{n}\sum_{i=1}^{n}(y_i - \hat{y}_i)^2} \quad (18)$$

$$\text{MAE} = \frac{1}{n}\sum_{i=1}^{n}|y_i - \hat{y}_i| \quad (19)$$

$$\text{Std} = \sqrt{\frac{1}{n}\sum_{i=1}^{n}(y_i - \mu)^2} \quad (20)$$

where $y_i$ denotes the ground-truth depth at sample $i$, $\hat{y}_i$ the corresponding predicted depth, $n$ the total number of samples, and $\mu$ the mean of the ground-truth depths. Lower values of RMSE, MAE, and Std indicate better prediction accuracy. For the classification task, OA and mIoU were calculated as:

$$\text{OA} = \frac{TP + TN}{TP + TN + FP + FN} \quad (21)$$

$$\text{mIoU} = \frac{1}{N}\sum_{i=1}^{N}\frac{TP_i}{TP_i + FP_i + FN_i} \quad (22)$$

where $TP, TN, FP$, and $FN$ denote the number of true positives, true negatives, false positives, and false negatives, respectively. $N$ is the number of classes, and $TP_i, FP_i$, and $FN_i$ are the true positives, false positives, and false negatives for class $i$. Higher OA and mIoU values indicate better classification performance.

## 4. Experimental results and analysis

### 4.1. Quantitative evaluation

Tables 2, 3, and 4 report the evaluation performance of the various models in the two study areas (Agia Napa and Puck Lagoon) using aerial, SPOT 6, and Sentinel-2 imagery.

#### 4.1.1. Comparison with empirical bathymetry models

The performance of Multiband Linear (Lyzenga, 1978) and Log Band Ratio (Stumpf et al., 2003) empirical models was consistently outperformed by Seabed-Net across all datasets and locations in terms of RMSE, MAE, and standard deviation. Seabed-Net achieves substantial reductions in RMSE, typically ranging from 40% to 60% for aerial data, 60% to 70% for SPOT 6, and 65% to 75% for Sentinel-2 imagery, depending on the location. The performance gap is especially pronounced in Agia Napa, a site with more complex bathymetric features and diverse seabed textures. In this area, the empirical models yield significantly higher errors, up to 2-3 times more than in the relatively uniform Puck Lagoon, indicating their limited capacity to generalize to challenging environments. In contrast, Seabed-Net maintains consistently low error across both sites. Its multitask architecture enables it to adapt well to heterogeneous seabed conditions and varying optical properties, where empirical models fail to cope with non-linearities and spatial variability.

#### 4.1.2. Comparison with machine learning regressors

Compared to Random Forest Regression and Support Vector Regression models, Seabed-Net consistently delivers superior performance across all sensors and locations. While Random Forest Regression and SVR exhibit moderate predictive capabilities with improved performance over empirical models, they still fall significantly short of Seabed-Net in terms of RMSE, MAE, and standard deviation. For instance, in the Agia Napa region, Seabed-Net reduces RMSE by approximately 24% to 42% compared to Random Forest, and by 32% to 49% compared to SVR, across all data sources. In the more uniform Puck Lagoon, Seabed-Net also maintains lower errors, outperforming Random Forest and SVR by margins of 14% to 39%. Notably, Seabed-Net shows much lower standard deviation values, reflecting its greater prediction stability. These improvements are most significant in Sentinel-2 imagery, where Seabed-Net achieves nearly 0.43 m RMSE in Puck Lagoon versus 0.642 m and 0.637 m from Random Forest and SVR respectively. The consistent gains of Seabed-Net highlight the advantages of deep multitask architectures that can capture complex spatial-spectral relationships and model non-linearities more effectively than conventional empirical and machine learning methods.

### 4.1.3. Comparison with single-task bathymetry model

Across all sensors and study areas, the proposed Seabed-Net consistently outperforms UNet-bathy in terms of RMSE, MAE, and standard deviation. Specifically, Seabed-Net achieves reductions in RMSE of approximately 10-22% for aerial data, 9-19% for SPOT 6, and 24-38% for Sentinel-2 across the two locations. MAE improvements are similarly substantial, ranging from 12-26% for aerial, 8-39% for SPOT 6, and 27-35% for Sentinel-2. In terms of standard deviation, reductions span 10-23% (aerial), 6-44% (SPOT 6), and 25-51% (Sentinel-2), demonstrating the model's superior consistency in depth predictions. This comparison highlights that the proposed multi-task model is not only effective at higher resolutions but is also more robust to resolution degradation, maintaining a consistent advantage over direct single-task bathymetric predictors across all settings.

### 4.1.4. Comparison with single-task pixel-based classification models

While both UNet and SegFormer achieve competitive OA, Seabed-Net consistently surpasses them across all sensors and locations. For aerial imagery, Seabed-Net improves OA by up to 1.67% compared to SegFormer. On SPOT 6 data, the improvement reaches up to 5.59%, and for Sentinel-2, the margin increases to as much as 8.72%. In addition to OA, Seabed-Net achieves higher mIoU scores across all data sources. For aerial imagery, SegFormer attains an mIoU of 0.846, while UNet lags behind at 0.820. On SPOT 6 data, SegFormer reaches 0.708 mIoU compared to UNet's 0.676. Sentinel-2 results show a similar trend, with SegFormer achieving 0.710 and UNet 0.643. These results highlight the superior segmentation performance of SegFormer over UNet, yet both are outperformed by our multi-task Seabed-Net in terms of both mIoU and OA.

Crucially, our proposed multi-task model maintains high OA and mIoU while simultaneously generating accurate, continuous bathymetry estimates. This dual capability underscores the limitations of pixel-based classification-only architectures in operational coastal mapping scenarios, where both classification and depth information are essential.

### 4.1.5. Comparison with multi-task learning models

The final comparison is against state-of-the-art multi-task learning approaches. Our proposed method consistently outperforms all compared multi-task models across both bathymetric estimation and seabed classification. In terms of bathymetry, Seabed-Net achieves relative RMSE reductions of up to 20% compared to MTL, approximately 10-15% compared to JSH-Net, and 7-15% compared to TaskPrompter across both study sites in the aerial data. Compared to PAD-Net and MTI-Net, which employ significantly larger architectures (81M and 128M parameters, respectively), Seabed-Net still delivers up to 30% lower RMSEs in some settings, indicating more efficient and accurate depth estimation. Even when compared to TaskPrompter which is the largest model in the cohort at 392M parameters, Seabed-Net consistently achieves lower RMSEs across the imagery from all sensors, with improvements of up to 12% on Sentinel-2 data.

For seabed classification, Seabed-Net also achieves higher OA than all other models. On the aerial data, it outperforms TaskPrompter and MTL by approximately 2% and JSH-Net by 4%. Gains are even more pronounced on SPOT 6 and Sentinel-2, where Seabed-Net exceeds TaskPrompter by over 5-7% in OA, PAD-Net by over 4-6%, and MTI-Net by up to 10%. In addition to OA, Seabed-Net demonstrates superior segmentation performance in terms of mIoU. On aerial imagery, Seabed-Net achieves an mIoU of 0.961, compared to 0.912 for TaskPrompter and 0.923 for JSH-Net. For SPOT 6 data, Seabed-Net obtains 0.784 mIoU, outperforming TaskPrompter (0.673), PAD-Net (0.631), and MTL (0.659). On Sentinel-2, Seabed-Net again leads with an mIoU of 0.742, versus 0.692 for TaskPrompter and 0.681 for MTL. The lower mIoU values in Agia Napa compared to Puck Lagoon are largely due to the increased number and complexity of seabed classes. Agia Napa includes four classes, such as macroalgae which is more difficult to distinguish and often occur in fragmented, spectrally ambiguous patches. In particular, macroalgae covers small areas that fall below the ground sampling distance (GSD) of SPOT 6 and Sentinel-2, leading to mixed pixels and higher misclassification rates. While OA remains high, mIoU is more sensitive to these per-class errors, especially in heterogeneous, multi-class settings like Agia Napa.

These results demonstrate Seabed-Net's superior ability to jointly learn bathymetry and seabed class representations, even under remote sensing sensor shifts and increased input variability. The model consistently generalizes well across both tasks, highlighting the advantages of its tailored multi-task architecture over parameter-heavy or pixel-based classification-biased baselines. Overall, Seabed-Net sets a new benchmark for bathymetry estimation within multi-task learning frameworks, achieving a strong balance between parameter efficiency, predictive accuracy, and robustness to spatial resolution degradation.

We think that the superior performance of Seabed-Net over single-task and prior multi-task baselines arises from its ability to jointly leverage depth and habitat information at multiple scales. The dual-branch encoders allow each task to learn specialized features, while the Attention Feature Fusion module injects localized pixel-based classification features into the depth branch (and vice versa), sharpening fine structural details. Meanwhile, the windowed Swin-Transformer Fusion captures long-range dependencies

**Table 2**

Average$_3$ testing performance obtained using the aerial data. Entities in bold indicate the best score.

| | Model | Params. (M) | Agia Napa [Aerial] | | | | | Puck Lagoon [Aerial] | | | | |
|---|---|---|---|---|---|---|---|---|---|---|---|---|
| | | | RMSE ↓ (m) | MAE ↓ (m) | Std. ↓ (m) | mIoU ↑ (%) | OA ↑ (%) | RMSE ↓ (m) | MAE ↓ (m) | Std. ↓ (m) | mIoU ↑ (%) | OA ↑ (%) |
| Emp. | Multiband Linear (Lyzenga, 1978) | - | 1.321 | 1.054 | 1.362 | - | - | 0.618 | 0.441 | 0.611 | - | - |
| | Log Band Ratio (Stumpf et al., 2003) | - | 1.613 | 1.598 | 1.689 | - | - | 0.543 | 0.412 | 0.512 | - | - |
| Regr. | Random Forest Regression | - | 0.862 | 0.612 | 0.853 | - | - | 0.378 | 0.247 | 0.378 | - | - |
| | Support Vector Regression | - | 1.007 | 0.699 | 0.959 | - | - | 0.412 | 0.269 | 0.412 | - | - |
| Single Task | UNet-bathy (Agrafiotis et al., 2024) | 1.9 | 0.616 | 0.420 | 0.598 | - | - | 0.298 | 0.187 | 0.297 | - | - |
| | UNet (Ronneberger et al., 2015) | 31 | - | - | - | 0.879 | 94.67 | - | - | - | 0.820 | 95.33 |
| | SegFormer (Xie et al., 2021) | 84.7 | - | - | - | 0.878 | 94.67 | - | - | - | 0.846 | 97.33 |
| Multi Task | PAD-Net (Xu et al., 2018) | 81 | 0.800 | 0.588 | 0.706 | 0.716 | 86.00 | 0.550 | 0.392 | 0.561 | 0.783 | 88.21 |
| | MTI-Net (Vandenhende et al., 2020) | 128 | 0.783 | 0.569 | 0.785 | 0.814 | 91.65 | 0.647 | 0.475 | 0.552 | 0.862 | 92.68 |
| | MTL (Ekim and Sertel, 2021) | 15.4 | 0.693 | 0.465 | 0.688 | 0.873 | 93.33 | 0.552 | 0.385 | 0.452 | 0.874 | 93.33 |
| | JSH-Net (Zhang et al., 2022a) | 21.6 | 0.621 | 0.458 | 0.615 | 0.833 | 91.33 | 0.440 | 0.268 | 0.434 | 0.923 | 96.00 |
| | TaskPrompter (Ye and Xu, 2024) | 392 | 0.596 | 0.482 | 0.589 | 0.869 | 94.67 | 0.442 | 0.278 | 0.439 | 0.912 | 95.33 |
| | Seabed-Net (ours) | 380 | **0.556** | **0.408** | **0.541** | **0.885** | **95.33** | **0.260** | **0.152** | **0.254** | **0.961** | **98.00** |

**Table 3**

Average$_3$ testing performance obtained using the SPOT 6 data. Entities in bold indicate the best score.

| | Model | Params. (M) | Agia Napa [SPOT 6] | | | | | Puck Lagoon [SPOT 6] | | | | |
|---|---|---|---|---|---|---|---|---|---|---|---|---|
| | | | RMSE ↓ (m) | MAE ↓ (m) | Std. ↓ (m) | mIoU ↑ (%) | OA ↑ (%) | RMSE ↓ (m) | MAE ↓ (m) | Std. ↓ (m) | mIoU ↑ (%) | OA ↑ (%) |
| Emp. | Multiband Linear (Lyzenga, 1978) | - | 2.221 | 1.623 | 2.143 | - | - | 0.842 | 0.543 | 0.839 | - | - |
| | Log Band Ratio (Stumpf et al., 2003) | - | 2.637 | 2.748 | 2.551 | - | - | 1.233 | 0.834 | 1.236 | - | - |
| Regr. | Random Forest Regression | - | 0.940 | 0.615 | 0.920 | - | - | 0.537 | 0.280 | 0.536 | - | - |
| | Support Vector Regression | - | 0.915 | 0.588 | 0.886 | - | - | 0.612 | 0.309 | 0.608 | - | - |
| Single Task | UNet-bathy (Agrafiotis et al., 2024) | 1.9 | 0.718 | 0.483 | 0.691 | - | - | 0.817 | 0.412 | 0.815 | - | - |
| | UNet (Ronneberger et al., 2015) | 31 | - | - | - | 0.519 | 76.00 | - | - | - | 0.676 | 85.33 |
| | SegFormer (Xie et al., 2021) | 84.7 | - | - | - | 0.566 | 77.33 | - | - | - | 0.708 | 86.66 |
| Multi Task | PAD-Net (Xu et al., 2018) | 81 | 1.172 | 0.842 | 1.030 | 0.614 | 78.00 | 0.675 | 0.361 | 0.668 | 0.631 | 77.85 |
| | MTI-Net (Vandenhende et al., 2020) | 128 | 0.775 | 0.567 | 0.753 | 0.525 | 73.33 | 0.619 | 0.368 | 0.586 | 0.637 | 77.85 |
| | MTL (Ekim and Sertel, 2021) | 15.4 | 0.679 | 0.464 | 0.670 | 0.524 | 72.67 | 0.532 | 0.303 | 0.532 | 0.659 | 79.87 |
| | JSH-Net (Zhang et al., 2022a) | 21.6 | 1.160 | 0.808 | 1.099 | 0.634 | 80.27 | 0.535 | 0.330 | 0.505 | 0.680 | 81.21 |
| | TaskPrompter Ye and Xu (2024) | 392 | 0.712 | 0.502 | 0.701 | 0.562 | 75.65 | 0.592 | 0.338 | 0.577 | 0.673 | 79.98 |
| | Seabed-Net (ours) | 380 | **0.651** | **0.445** | **0.649** | **0.645** | **82.00** | **0.457** | **0.254** | **0.457** | **0.784** | **87.92** |

**Table 4**

Average$_3$ testing performance obtained using the Sentinel-2 data. Entities in bold indicate the best score.

| | Model | Params. (M) | Agia Napa [Sentinel-2] | | | | | Puck Lagoon [Sentinel-2] | | | | |
|---|---|---|---|---|---|---|---|---|---|---|---|---|
| | | | RMSE ↓ (m) | MAE ↓ (m) | Std. ↓ (m) | mIoU ↑ (%) | OA ↑ (%) | RMSE ↓ (m) | MAE ↓ (m) | Std. ↓ (m) | mIoU ↑ (%) | OA ↑ (%) |
| Emp. | Multiband Linear (Lyzenga, 1978) | - | 2.412 | 1.824 | 2.395 | - | - | 1.732 | 1.341 | 1.723 | - | - |
| | Log Band Ratio (Stumpf et al., 2003) | - | 0.917 | 0.724 | 0.913 | - | - | 0.937 | 0.639 | 0.941 | - | - |
| Regr. | Random Forest Regression | - | 1.239 | 0.783 | 1.203 | - | - | 0.670 | 0.393 | 0.669 | - | - |
| | Support Vector Regression | - | 1.137 | 0.747 | 1.079 | - | - | 0.642 | 0.386 | 0.637 | - | - |
| Single Task | UNet-bathy (Agrafiotis et al., 2024) | 1.9 | 1.068 | 0.694 | 0.940 | - | - | 0.907 | 0.493 | 0.874 | - | - |
| | UNet (Ronneberger et al., 2015) | 31 | - | - | - | 0.571 | 78.76 | - | - | - | 0.643 | 78.52 |
| | SegFormer (Xie et al., 2021) | 84.7 | - | - | - | 0.578 | 75.51 | - | - | - | 0.710 | 82.55 |
| Multi Task | PAD-Net (Xu et al., 2018) | 81 | 2.665 | 1.387 | 2.468 | 0.518 | 73.47 | 0.615 | 0.395 | 0.601 | 0.543 | 71.81 |
| | MTI-Net (Vandenhende et al., 2020) | 128 | 1.387 | 1.148 | 1.284 | 0.536 | 71.42 | 0.897 | 0.656 | 0.740 | 0.579 | 74.50 |
| | MTL (Ekim and Sertel, 2021) | 15.4 | 0.789 | 0.550 | 0.788 | 0.623 | 78.91 | 0.541 | 0.284 | 0.541 | 0.681 | 81.21 |
| | JSH-Net (Zhang et al., 2022a) | 21.6 | 0.856 | 0.609 | 0.845 | 0.626 | 79.59 | 0.480 | 0.279 | 0.479 | 0.662 | 79.87 |
| | TaskPrompter (Ye and Xu, 2024) | 392 | 0.753 | 0.521 | 0.748 | 0.624 | 79.99 | 0.502 | 0.256 | 0.501 | 0.692 | 81.95 |
| | Seabed-Net (ours) | 380 | **0.660** | **0.451** | **0.658** | **0.643** | **84.35** | **0.431** | **0.253** | **0.429** | **0.742** | **85.23** |

across the entire scene, preserving global context without sacrificing spatial resolution. Finally, dynamic uncertainty weighting prevents one task from dominating the shared representations, ensuring stable convergence and mutual reinforcement: depth features help delineate habitat boundaries, and semantic labels clarify ambiguous bottom reflectance for more accurate bathymetry. Collectively, these components enable Seabed-Net to precisely classify seabed structures and resolve depth in low-contrast regions, yielding marked improvements in both tasks.

### 4.2. CATZOC evaluation of the predicted depths

In this subsection, we evaluate the predicted bathymetry using the CATZOC (S-57/S-101) criteria (IHO, 2023). The Category Zone of Confidence (CATZOC) framework, established by the International Hydrographic Organization (IHO), provides a standardized approach to assess the quality and reliability of bathymetric data. By assigning a horizontal and vertical CATZOC rating, we assess the suitability of predicted depths for applications such as navigation, habitat mapping, and coastal monitoring.

Regarding the horizontal uncertainty thresholds, the value for CATZOC A1, for example, is calculated by combining a maximum allowable horizontal position error of 5 meters with an additional uncertainty of 5% of the depth. Sentinel-2 L2A orthorectified imagery, with a CE95 of approximately 11 m (Clerc, 2018), exceeds the A1 threshold for all depths up to 20 m but remains within the A2 limit. Consequently, Sentinel-2 data are classified as ZOC A2. Similarly, SPOT 6 orthoproducts, exhibiting a CE90 of 6-10 m (Airbus Defence and Space, 2025) surpass the A1 tolerance beyond the very shallow zone and therefore fall into ZOC A2 for typical shallow-water applications. In contrast, aerial orthomosaics demonstrate planar accuracies of 0.01 to 0.05 m (Agrafiotis et al., 2024), satisfying the ZOC A1 criteria throughout the entire 0-20 m depth range.

Regarding vertical uncertainty thresholds these thresholds are specified using a total vertical uncertainty model (TVU) that accounts for both a constant error term and a proportional depth term. Each CATZOC category (e.g., A1, A2, B, C, D) prescribes specific limits on vertical uncertainty, with higher-quality zones (such as A1) permitting tighter bounds-for example, ±0.50 m + 1% of depth. Results in Agia Napa are consistent across aerial, SPOT 6, and Sentinel-2 bathymetry: in all three cases, 57.14% of the depths meet A1 standards, and 42.86% are classified as A2, with no predictions falling into lower-quality categories C or D. In Puck Lagoon, performance shows stronger variability across resolutions. Using Sentinel-2 imagery, 68.00% of predictions fall within A1, 21.00% within A2, 8.00% within C, and 8.00% within D. With SPOT 6, 65.00% are classified as A1, 20.00% as A2, 9.00% as C, and 9.00% as D. Aerial imagery achieves 64.00% A1 coverage and 23.00% A2, but also includes 11.00% in class C and 11.00% in class D.

Despite these variations, the majority of the bathymetric predictions across all settings meet the accuracy thresholds for high-quality survey zones (A1 and A2). Moreover, in terms of data completeness, the aerial-derived bathymetry for both sites fulfills the CATZOC A1 requirement of 100% seafloor coverage. While feature detection at the 2 m scale was not directly evaluated, the imagery's GSD of 0.25 m suggests that such features would be visually resolvable within the data.

### 4.3. Qualitative evaluation

Beyond quantitative metrics, visual analysis provides critical insights into model behavior, especially in complex coastal environments where annotated data are sparse. To this end, we qualitatively assess the outputs of Seabed-Net against other deep learning-based baseline models by examining the predicted seabed classes and bathymetric maps over extended spatial extents. We structure this section into two parts, focusing separately on bathymetry estimation and seabed pixel-based classification.

#### 4.3.1. Bathymetry

Visual inspection of the derived bathymetry (Fig. 5 and Fig. 6) confirms the Seabed-Net's ability to capture fine-scale topographic variation in close correspondence with known seabed substrate and habitat types across multiple remote sensing sensors. Across both Agia Napa and Puck Lagoon, Seabed-Net (column h) consistently produces depth surfaces with clearer structural delineation compared to baseline models.

In Agia Napa (Fig. 5, seagrass-covered regions are distinctly outlined and identified as slight depth elevations relative to adjacent sandy areas, maintaining well-preserved boundary transitions. These elevated patches appear consistently across sensors (aerial, SPOT 6, Sentinel-2) when predicted by Seabed-Net, while other models often show fragmentation or blurred transitions. Sandy areas are rendered with broad, gradual slopes with scattered subtle undulations, especially evident in the higher-resolution aerial imagery. In contrast, rocky areas and reef zones produce abrupt depth changes are seen as sharp gradients and ridges, captured most accurately in Seabed-Net and TaskPrompter outputs, while earlier models (e.g., UNet-Bathy, PAD-Net) tend to under-represent or oversmooth these features. Macroalgal covered areas, especially in Sentinel-2 imagery, show irregular patches and fine canopy relief where Seabed-Net preserves these spatially variant textures more effectively than competing methods.

In Puck Lagoon (Fig. 6), the performance gap becomes even more pronounced. Seabed-Net's outputs remain stable, with low noise and high spatial coherence. Competing methods frequently introduce speckling, artifacts, or over-smoothing, particularly PAD-Net and JSH-Net, leading to loss of geomorphological detail. Notably, Seabed-Net retains finer-scale bathymetric structure in flat zones while avoiding the exaggerated patchiness observed in MTL-Net and MTI-Net results. This is particularly visible in the Sentinel-2 data, where Seabed-Net maintains depth consistency across larger patches without sacrificing local variation.

These qualitative differences reinforce the quantitative findings: Seabed-Net not only reduces RMSE but also yields topographically coherent and interpretable depth maps aligned with ecological and geological classes. This coherence is essential for downstream tasks like habitat classification, change detection, or navigational planning, where spatial fidelity and structural continuity are as important as absolute accuracy.

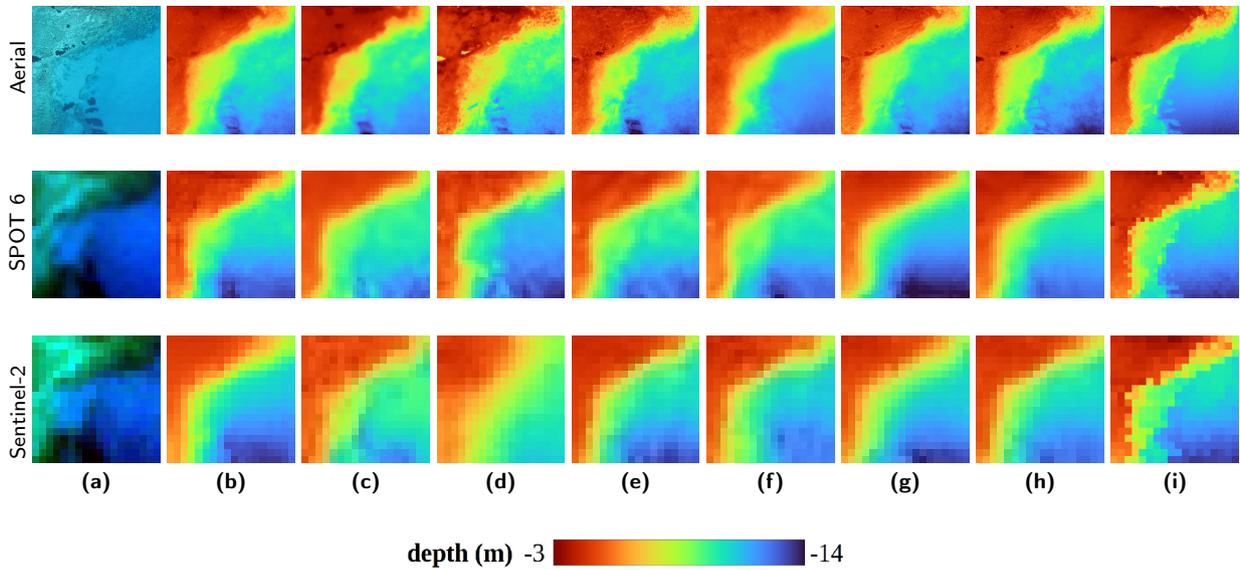

**Figure 5:** Bathymetry retrieval results on the aerial, SPOT 6 and Sentinel-2 imagery from the compared single-task and multi-task approaches. (a) True color composite of example patches acquired over Agia Napa, bathymetry obtained by (b) UNet-Bathy, (c) PAD-Net, (d) MTI-Net, (e) MTL, (f) JSH-Net, (g) TaskPrompter, (h) Seabed-Net and (i) LiDAR/SONAR.

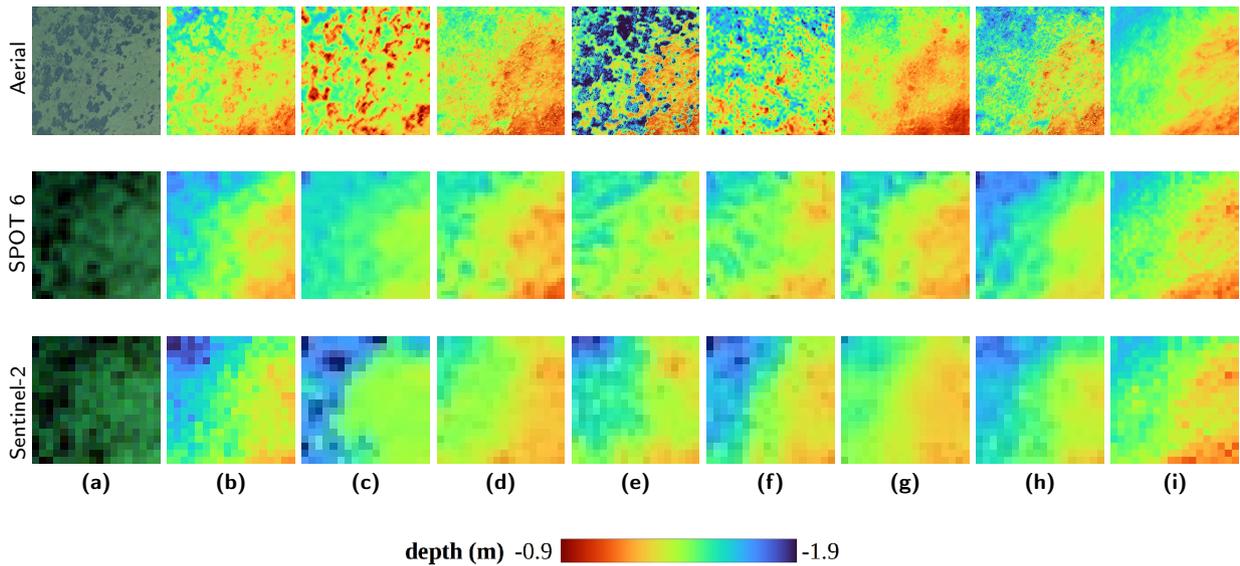

**Figure 6:** Bathymetry retrieval results on the aerial, SPOT 6 and Sentinel-2 imagery from the compared single-task and multi-task approaches. (a) True color composite of example patches acquired over Puck Lagoon, bathymetry obtained by (b) UNet-Bathy, (c) PAD-Net, (d) MTI-Net, (e) MTL, (f) JSH-Net, (g) TaskPrompter, (h) Seabed-Net and (i) LiDAR/SONAR.

### 4.3.2. Pixel-based classification

Visual inspection of predicted seabed class maps in Fig. 7 and Fig. 8, covering Agia Napa and Puck Lagoon respectively, shows that Seabed-Net delivers consistently superior spatial coherence and semantic accuracy compared to baseline methods. Across the used imagery, competing models often exhibit visible artifacts such as blockiness, edge inconsistency, and irregular fragmentation, especially evident in lower-resolution inputs. These issues frequently lead to habitat boundaries that are either overly smoothed or irregular with noisy boundaries, distorting the natural structure of benthic zones.

In contrast, Seabed-Net (column i) preserves habitat morphology with notably higher fidelity. In Agia Napa (Fig. 7), seagrass blobs appear as spatially coherent patches that follow their natural contours, avoiding the artificial clustering seen in PAD-Net, MTI-Net, or JSH-Net. Sandy regions are accurately depicted with minimal noise, while rocky zones are sharply defined, particularly in the aerial and SPOT 6 imagery. Macroalgal habitats, which tend to be confused with seagrass in other models due to their spectral similarity and sparse supervision, are better localized and appropriately fragmented, aligning with their ecologically patchy nature.

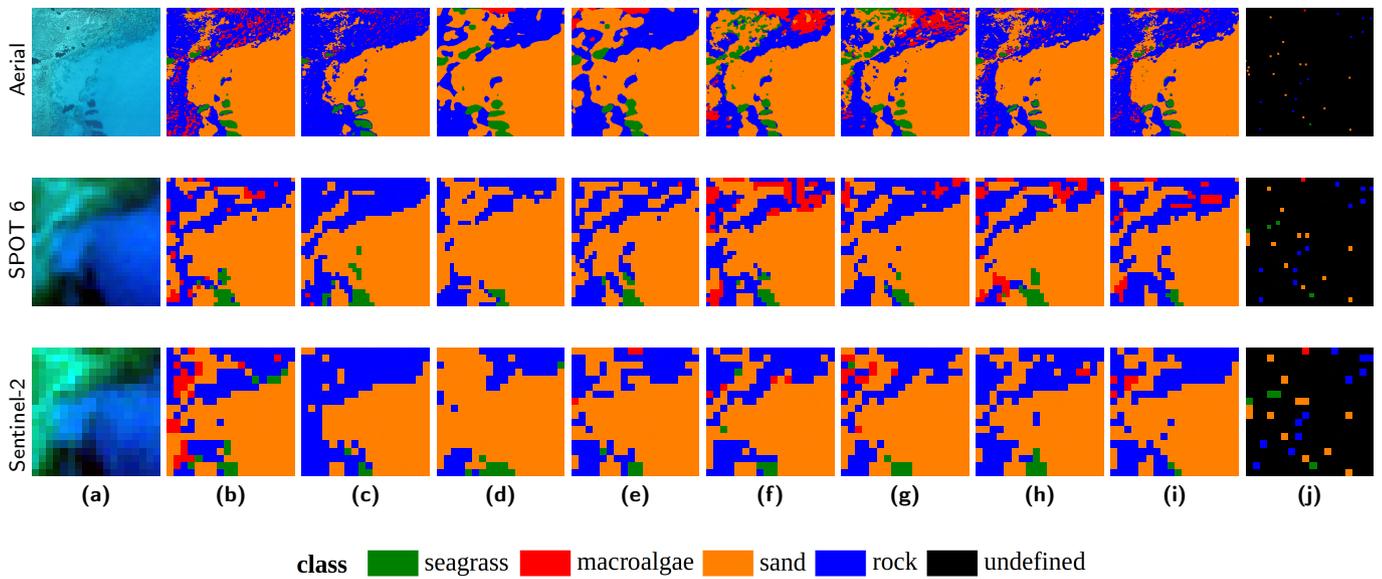

**Figure 7:** Pixel-based classification results on the aerial, SPOT 6 and Sentinel-2 imagery from the compared single-task and multi-task approaches. (a) True color composite of example patches acquired over Agia Napa, seabed classes obtained by (b) U-Net, (c) SegFormer, (d) PAD-Net, (e) MTI-Net, (f) MTL, (g) JSH-Net, (h) TaskPrompter, (i) Seabed-Net, and (j) seabed reference samples.

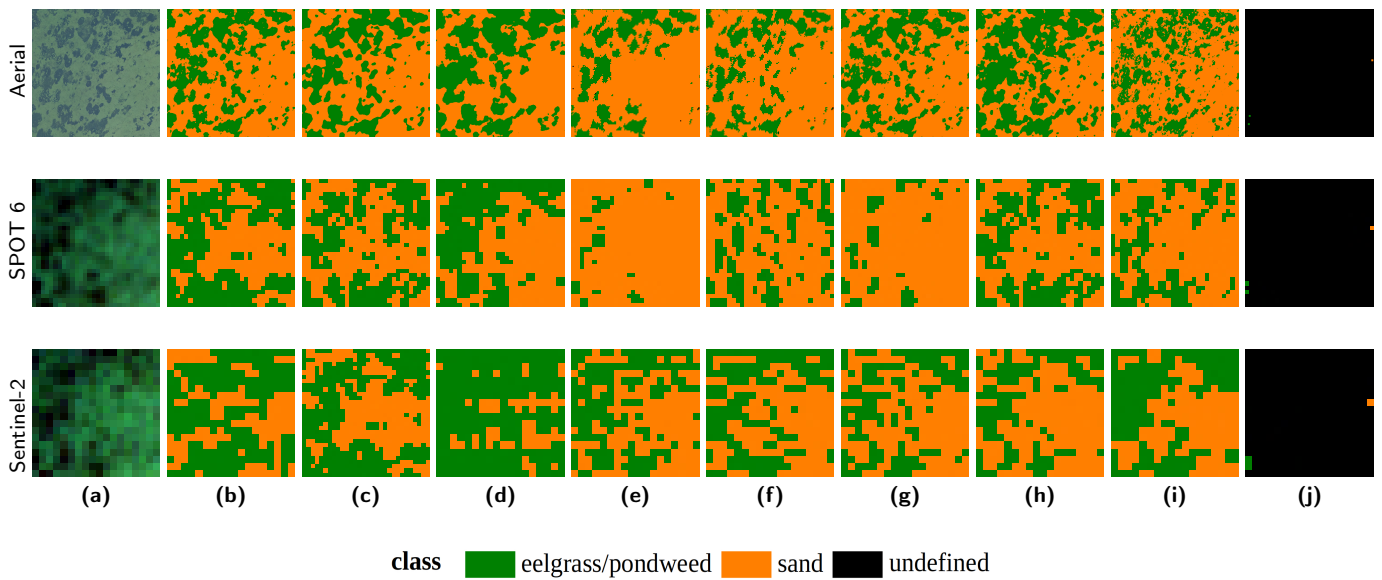

**Figure 8:** Pixel-based classification results on the aerial, SPOT 6 and Sentinel-2 imagery from the compared single-task and multi-task approaches. (a) True color composite of example patches acquired over Puck Lagoon, seabed classes obtained by (b) U-Net, (c) SegFormer, (d) PAD-Net, (e) MTI-Net, (f) MTL, (g) JSH-Net, (h) TaskPrompter, (i) Seabed-Net, and (j) seabed reference samples.

In Puck Lagoon (Fig. 8), where only two dominant classes (eelgrass/pondweed and sand) are present and optical contrast is lower, the differences become even more significant. Baselines like PAD-Net and MTL often result in heavy class imbalance or random speckling, particularly in SPOT 6 and Sentinel-2 imagery. Seabed-Net, however, achieves more stable and ecologically plausible segmentations: eelgrass/pondweed patches are separated from the surrounding sand flats, avoiding both excessive smoothing and inconsistent misclassification.

These qualitative improvements reflect the benefits of joint optimization. By learning depth and pixel-based classification together, Seabed-Net leverages bathymetric features to inform class boundary precision, especially in optically ambiguous areas where single-task classifiers often fail. For

P. Agrafiotis and B. Demir: *Revised article* Page 13 of 21

instance, darker deep zones which are commonly misclassified as vegetated areas due to intensity-based confusion, are more accurately labeled, as depth prediction informs the network that these are not biologically dense shallows.

This cross-task synergy results in higher structural consistency and better pixel-based classification, especially under low supervision and challenging optical conditions. The output is a set of seabed maps that are not only more accurate in terms of class labels but also better aligned with the underlying seabed structure and ecological organization of shallow-water environments.

### 4.4. Accuracy analysis by depth intervals and seabed habitats

In this subsection, we perform a finer-scale quantitative evaluation of Seabed-Net's bathymetric and pixel-based classification accuracy by segmenting results across different depth intervals and diverse seabed types. To that direction, a detailed breakdown by depth ranges and habitat types offers a deeper insight into the stability and reliability of the proposed model in varying environmental conditions. This stratified approach reveals how consistently the model performs under different environmental conditions, highlighting its strengths and exposing any limitations.

*4.4.1. Bathymetry retrieval*

Fig. 9 presents the bathymetry retrieval performance across seabed classes and depth ranges in Agia Napa and Puck Lagoon, using aerial, SPOT 6, and Sentinel-2 imagery. Across both sites, aerial data consistently yields the lowest RMSE values, particularly in shallow waters (0-4 m), where errors generally remain below 0.5 m. SPOT 6 and Sentinel-2 exhibit increased RMSE, with Sentinel-2 showing particularly high errors exceeding 2.0 m in the 4-6 m range in Puck Lagoon and the 8-10 m range in Agia Napa for certain classes. These results are consistent with the environmental and observational constraints present during image acquisition. Specifically, in Puck Lagoon, Secchi depth measurements ranged from 3.5m to 3.8m, indicating low water clarity and thus limited effective optical penetration. This makes reliable depth retrieval beyond 3.8m inherently difficult. In contrast, Agia Napa exhibited much clearer waters, with Secchi depths between 22.2m and 26.1m, allowing for greater depth visibility. Seagrass and sand substrates generally produce lower RMSEs than rock or macroalgae, likely due to their more consistent spectral reflectance. Regarding macroalgae, elevated RMSE values in SPOT 6 and Sentinel-2 imagery are partly attributed to the small spatial extent of macroalgae patches, which often fall below the pixel size of these sensors, leading to mixed-pixel effects. In the 4-6 m depth range, both SPOT 6 and Sentinel-2 show markedly higher errors in Puck Lagoon compared to Agia Napa. This discrepancy is due to higher concentrations of chlorophyll in Puck Lagoon, which reduce water clarity and negatively affect optical depth retrieval, especially from space. This underscores the combined effect of sensor resolution, water visibility, and bottom type on bathymetric accuracy, as well as the inherent data limitations in training and validation. Missing bars in several depth-class bins indicate the absence of some seabed classes at specific depth intervals.

Interestingly, in Agia Napa, a decrease in RMSE beyond 10 m is observed for the "sand" and "all" categories. At these greater depths, the seabed is predominantly composed of sandy bottom, which is optically simpler and exhibits spectrally homogeneous reflectance characteristics. This contributes to a lower RMSE, as such uniform substrates are easier for any model to interpret consistently. Moreover, these sandy areas typically lack complex features such as abrupt topographic changes, further reducing variability in depth predictions.

*4.4.2. Pixel-based classification*

The classification performance, as shown in Fig. 10, varies considerably with sensor type, water depth, and seabed composition. In both Agia Napa and Puck Lagoon, aerial imagery provides the most stable and accurate classification across all depth ranges, benefiting from higher spatial resolution and clearer separation between seabed classes. In contrast, SPOT 6 maintains moderate accuracy but shows noticeable deterioration beyond 4 m depth, while Sentinel-2 struggles especially in deeper waters, with increased misclassification rates. This decline in classification accuracy with depth is primarily due to reduced light penetration, increased water column distortion, and a weakening of spectral separability between benthic classes. Lower-resolution sensors like SPOT 6 and Sentinel-2 are also more affected by these depth-related challenges, as smaller seabed features become harder to resolve and discriminate. Seagrass and sand are generally better distinguished than macroalgae and rock, which tend to have more spectral overlap and less distinct texture. The reduced per-class accuracy values for macroalgae in SPOT 6 and Sentinel-2 imagery are partly attributed to the limited spatial extent of macroalgae patches, which often fall below the sensors' pixel sizes. The absence of bars for specific classes in certain depth intervals reflects the lack of annotated pixels in those regions. In contrast to Fig. 9, which uses predicted classes across the entire depth range, the per-class accuracy metrics in Fig. 10 are computed only for annotated pixels. In the case of Puck Lagoon, annotations were limited to the 0–4 m depth range due to restricted water visibility, consistent with the previously reported Secchi depths of 3.5–3.8 m during image acquisition. As a result, classification accuracy values are not available for deeper depths (4-8 m) in this region. These findings underscore the importance of high-resolution imagery when mapping heterogeneous and spatially fine-scale benthic habitats, especially in deeper or more optically complex waters.

### 4.5. Ablation study

To better understand the contribution of each architectural component to the overall performance of Seabed-Net, we conducted a detailed ablation study on the Agia Napa Sentinel-2 dataset. Specifically, we examined the impact of the Feature Alignment and Aggregation (FAA) modules and

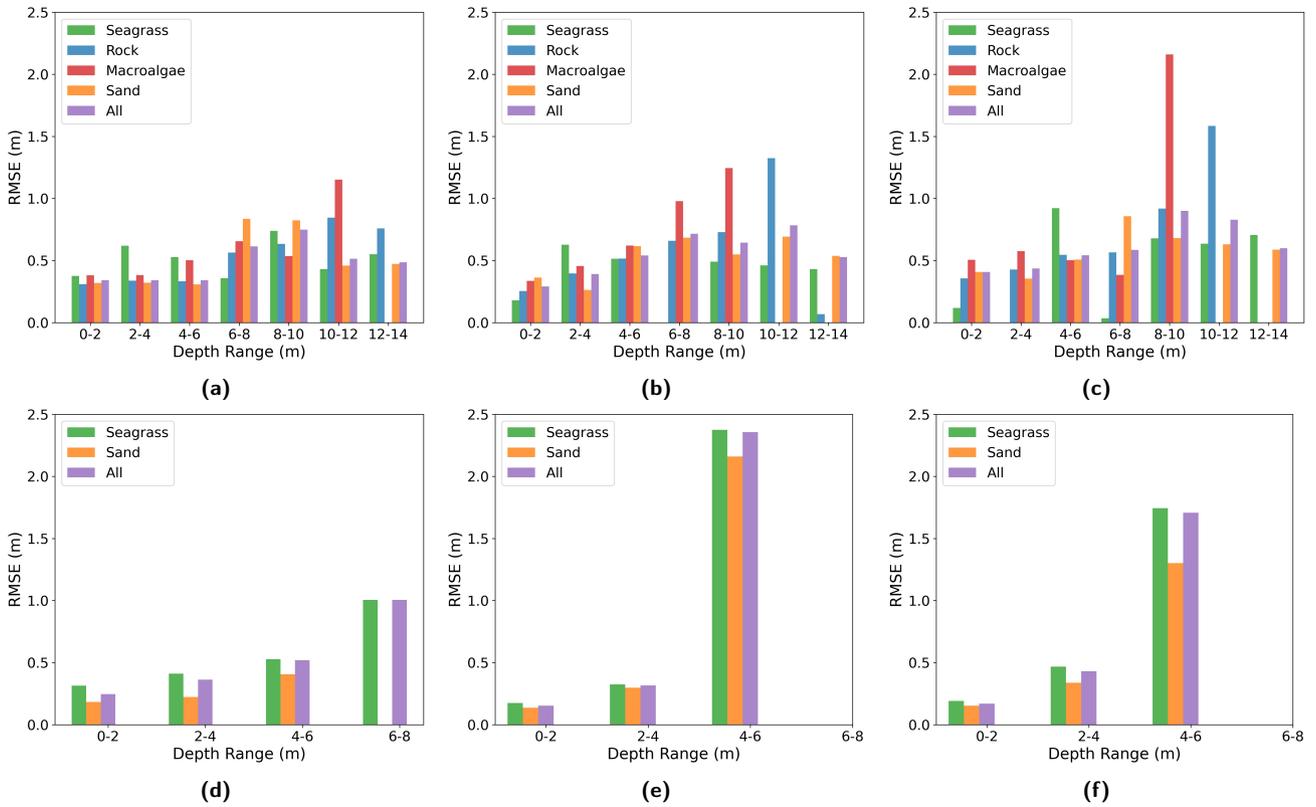

**Figure 9:** Bathymetric RMSE by depth range and seabed class in Agia Napa for (a) aerial, (b) SPOT 6, and (c) Sentinel-2 data. Corresponding metrics in Puck Lagoon for (d) aerial, (e) SPOT 6, and (f) Sentinel-2 data.

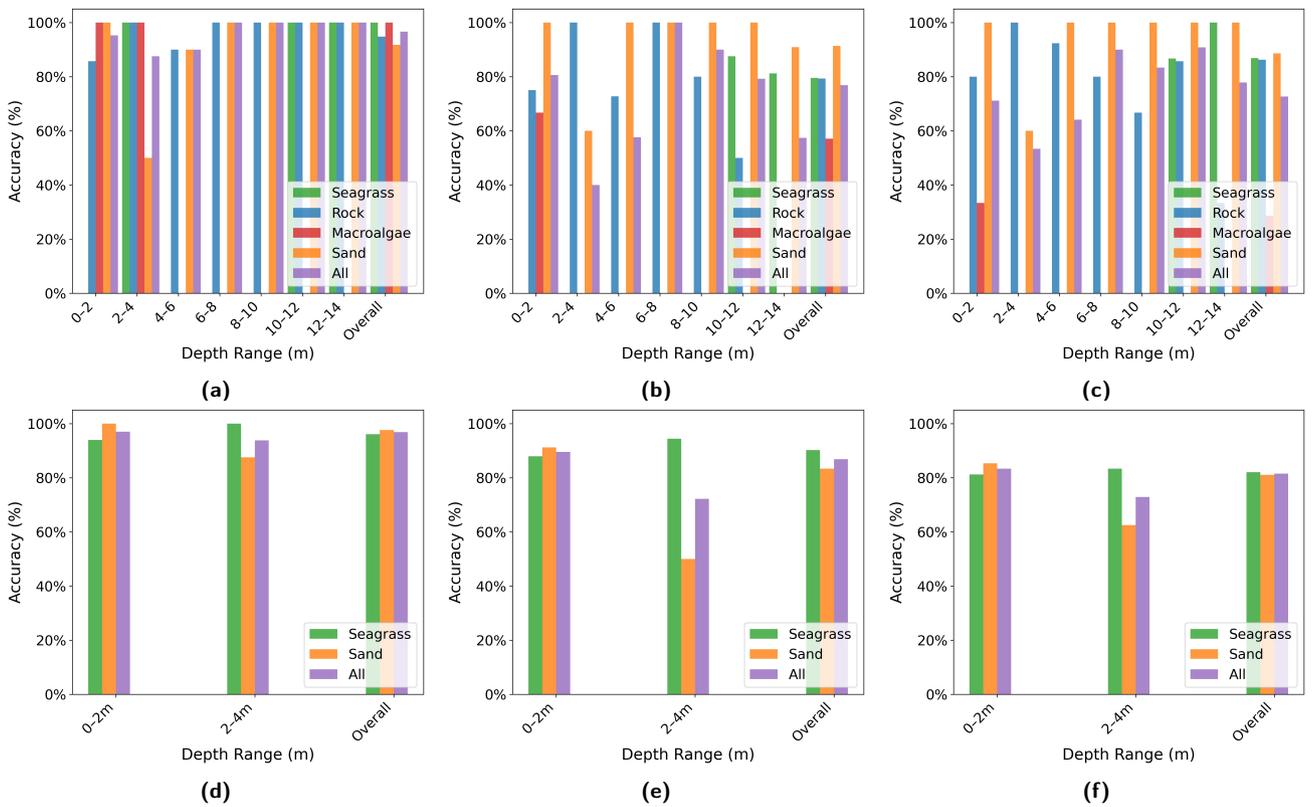

**Figure 10:** Per-class accuracy of pixel-based classification by depth range and seabed class in Agia Napa for (a) aerial, (b) SPOT 6, and (c) Sentinel-2 data. Corresponding metrics in Puck Lagoon for (d) aerial, (e) SPOT 6, and (f) Sentinel-2 data.

**Table 5**
Ablation study demonstrating the Average$_3$ testing performance on the Agia Napa Sentinel-2 data compared to the reference data.

| Setting | RMSE(m) | OA% | Params. |
|---|---|---|---|
| 4 FAA and 4 ViT blocks | **0.660** | **84.35** | 380M |
| 3 FAA and 3 ViT blocks | 0.891 | 76.19 | 106M |
| No FAA and ViT blocks | 1.125 | 74.83 | 15M |
| No FAA blocks | 0.963 | 78.91 | 379M |
| No ViT blocks | 0.872 | 77.55 | 16M |
| No spatial attention in AFF | 0.923 | 80.27 | 380M |
| No channel attention in AFF | 1.012 | 80.95 | 380M |
| 2 ViT heads (instead of 4) | 0.854 | 76.87 | 380M |
| No cross attention in ViT | 0.873 | 82.31 | 379M |
| Bathymetry branch with BN | 1.567 | 79.59 | 379M |
| Classification branch without BN | 0.883 | 80.67 | 379M |

Vision Transformer (ViT) blocks, as well as their quantity, the presence or absence of attention mechanisms, and key hyperparameters such as attention heads and window size. The goal was to isolate and quantify the influence of each design choice on bathymetric accuracy (measured in RMSE), classification performance (OA%), and model complexity (parameter count).

The ablation study results shown in Table 5 highlight the importance of Seabed-Net's architectural components. The full model, with 4 FAA and 4 ViT blocks, achieves the best performance, while reducing or removing these blocks leads to consistent drops in both bathymetric accuracy and pixel-based classification performance. FAA modules have a stronger impact on depth estimation, while ViT blocks contribute more to pixel-based classification. Performance declines when attention mechanisms are simplified; particularly when channel attention or cross-attention is removed, demonstrating their role in enhancing feature fusion and task interaction. Performance also declines when batch normalization was added to the bathymetry branch or removed from the classification branch, confirming it's task-specific use in Seabed-Net. These results confirm that both the depth and design of the network are key to achieving robust multi-task performance.

## 5. Discussion

This work introduces Seabed-Net, a robust multi-task learning framework designed to jointly estimate bathymetry and perform pixel-based seabed classification from remote sensing imagery of various resolution. The proposed method demonstrates significant improvements over state-of-the-art single-task and multi-task baselines across multiple remote sensing sensors and environments. The results confirm the advantages of fusing complementary spatial, spectral, and semantic features in a unified architecture. However, like any remote sensing approach, performance limitations are imposed by the spatial resolution of the data.

### 5.1. Effect of spatial resolution on models performance

Spatial resolution emerged as a critical factor affecting both bathymetric accuracy and pixel-based classification performance. As expected, higher resolution aerial imagery enabled the most detailed representation of coastal features, producing the lowest RMSE and highest OA values. As resolution decreased, moving from aerial (0.25 m) to SPOT 6 (6 m) and Sentinel-2 (10 m) data, a significant performance degradation was observed in all the compared models. On average, the degradation from aerial to SPOT 6 data results in an increase of approximately 0.1-0.2 m in RMSE and a decrease of 10-15% in OA, depending on the model. Sentinel-2 imagery, with the lowest spatial resolution, intensifies these effects further, leading to the highest RMSEs and lowest OAs across all methods.

In particular, the increase in bathymetric error metrics from aerial to SPOT 6 and Sentinel-2 imagery is more pronounced for single-task methods. This underscores the advantage of multi-task learning: by sharing representations between tasks, Seabed-Net is able to maintain structural and contextual integrity in lower-resolution inputs. Depth features guide classification in ambiguous regions, while pixel-based classification reinforces bathymetric consistency in low-contrast or optically complex areas. The result is enhanced robustness to resolution degradation which is crucial for operational use in regions where only medium- or low-resolution imagery is available.

### 5.2. Task-dependent gains under varying spatial resolutions

The significantly larger improvements observed in bathymetric performance, highlight a substantial opportunity for targeted innovation in bathymetric modeling. This is particularly evident in Fig. 11, where RMSE improvements consistently outpace gains in overall accuracy (OA) across all sensor types. These improvements (Δ) are calculated as follows:

$$\Delta OA\% = \left( \frac{OA_{Ours} - OA_{Best\ MTL}}{OA_{Best\ MTL}} \right) \times 100 \quad (23)$$

$$\Delta RMSE\% = \left( \frac{RMSE_{Best\ MTL} - RMSE_{Ours}}{RMSE_{Best\ MTL}} \right) \times 100 \quad (24)$$

These gains are especially noticeable for lower-resolution sensors like Sentinel-2, where RMSE improvements reach over 14%, while OA improvements remain more modest (around 4-8%). While these values are not directly comparable due to the different nature and scale of the metrics (continuous RMSE vs. categorical OA), this contrast underscores two key observations.

First, it reinforces the idea that current bathymetric models, often adapted from terrestrial depth estimation techniques (e.g., from autonomous driving or land-based photogrammetry), are not well-optimized for aquatic remote sensing data. Marine environments introduce unique challenges such as sunglint, water surface variability, turbidity, seasonal variation, and spectral and geometric distortions due to water column effects. These are not present in land-based datasets. The superior RMSE gains suggest that multi-task learning or task-specific adaptations allow networks to begin capturing these aquatic-specific complexities more effectively.

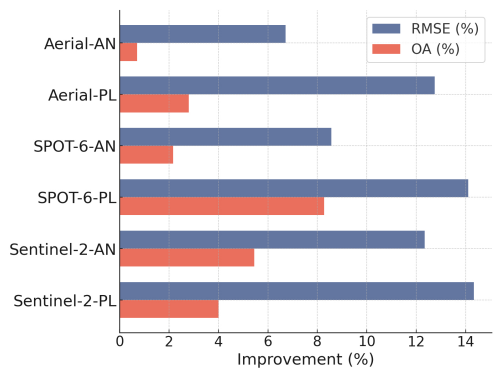

**Figure 11:** Seabed-Net's performance improvements over the best compared models for each remote sensing sensor and area. AN stands for Agia Napa while PL for Puck Lagoon.

Second, the trend of larger bathymetric improvements at lower resolutions (e.g., Sentinel-2 and SPOT 6) implies that multi-task learning and specifically Seabed-Net provides a particularly important boost in scenarios where spatial resolution is insufficient to distinguish fine-scale features. In such cases, the incorporation of auxiliary tasks or domain-specific constraints allows the network to compensate for the loss of spatial detail by leveraging cross-task regularities and shared representations. This is a crucial finding, as many available and widely-used satellite platforms fall into this lower-resolution category. The performance gap indicates that these platforms benefit most from smarter and more integrated model designs.

In contrast, the relatively modest improvements in OA for pixel-based classification across all sensors suggest that classification tasks may already be approaching a saturation point under current modeling approaches. The limited gains could reflect the fact that classification networks, especially when not explicitly adapted for aquatic features, are extracting close to the maximum discriminative information possible from the spectral and spatial input data without additional constraints or data types.

### 5.3. Dataset-specific observations

The environmental characteristics of the surveyed regions Agia Napa and Puck Lagoon, also influenced model performance. Puck Lagoon consistently produced lower overall RMSE and higher classification OA across all models and resolutions. This can be attributed to it's simpler bathymetric and substrate structure and smaller average and maximum depth. In contrast, the greater habitat diversity and variable bottom types in Agia Napa posed increased challenges, especially under low-resolution imagery. These findings suggest that regional variability must be considered in model deployment.

It should also be acknowledged that while the training and testing data were strictly non-overlapping at the patch level, they were nonetheless derived from the same orthoimage or satellite scene. This implies that both training and testing patches share acquisition conditions such as sun/sensor geometry, atmospheric state, sea surface, and water quality. As such, although the results reflect generalization to spatially disjoint patches, they may still be optimistic relative to scenarios where models are applied to entirely independent images, even from the same geographical region. This limitation is common practice in remote sensing studies due to the scarcity of repeated acquisitions under various conditions, but is nevertheless important to recognize. Incorporating domain adaptation mechanisms or fine-tuning strategies could therefore further improve generalization to unseen or more complex coastal settings.

### 5.4. Mixed-Pixel effects

Mixed-pixel effects are a common limitation in all supervised methods for pixel-based classification, particularly when using medium- and low-resolution imagery such as SPOT 6 and Sentinel-2. In these settings, narrow or fragmented features, such as macroalgae patches, often occupy areas smaller than a single pixel, leading to pixels that contain a mixture of seabed types. Supervised models trained on discrete labels are then forced to assign these mixed pixels to the nearest pure class, introducing classification errors that can also propagate into depth estimation when using multitask models.

In this work, to mitigate label noise, we employ sparse annotations focused on confidently labeled, "pure" pixels. This reduces confusion during training but does not fully address the ambiguity encountered at inference time, where the model must still process mixed spectral signatures. While sparse labeling helps limit training bias, it inherently limits the model's exposure to the full variability of real-world conditions. Future improvements could include soft labeling strategies or probabilistic class maps, which may better account for class mixtures and reduce error propagation from classification to depth estimation.

### 5.5. Considerations on data splitting and spatial dependence

In this study, we employed a random splitting strategy to divide the dataset into training and testing subsets, in line with the protocol established in previous work on MagicBathyNet (Agrafiotis et al., 2024). While there was no overlap between our training and testing data, we acknowledge that random splitting in the presence of spatially autocorrelated data, such as remote sensing imagery used in shallow-water bathymetry, can lead to an overestimation of model performance. This occurs because calibration and test samples located in close spatial proximity may share similar environmental and spectral characteristics, resulting in information leakage between the two sets. As demonstrated in

**Table 6**

Inference time in seconds (s), number (#) of trainable parameters in millions (M), RMSE in meters (m) and OA for Agia Napa using the Sentinel-2 data.

| Metric | Seabed-Net | | MTL | |
| --- | --- | --- | --- | --- |
| | Depth | Classes | Depth | Classes |
| # of Parameters (M) | 380 | | 15.4 | |
| Inference Time (s) | 1.99 | 5.17 | 1.92 | 4.45 |
| RMSE (m) | 0.660 | - | 0.789 | - |
| OA (%) | - | 84.35 | - | 78.91 |

recent work (Knudby and Richardson, 2023), such leakage can disproportionately benefit models that are prone to overfitting, yielding optimistic performance estimates that may not generalize well to spatially disjoint or unseen regions.

Although random splits are common practice, especially when evaluating models in data-rich settings where many applications involve dense calibration inputs and local interpolation, we recognize that for generalization to new, unsurveyed areas, blocked spatial splitting may offer a more realistic assessment of a model's predictive capabilities. In future work, we intend to explore spatially disjoint validation protocols and provide predefined spatial splits for the MagicBathyNet (Agrafiotis et al., 2024) dataset to enable fairer benchmarking and a deeper understanding of spatial generalization limits in multi-modal SDB tasks.

### 5.6. Model complexity, inference time and real-time applicability

The number of parameters appears to play a crucial role in model performance, particularly when dealing with high variability in spatial resolution, as well as the inherent spectral and radiometric variability of aquatic environments. Factors such as changes in turbidity, chlorophyll concentration, water depth, surface roughness, sunglint, and bottom reflectance significantly impact the quality and consistency of remote sensing data. Our proposed model, Seabed-Net, with 380M parameters, is among the largest evaluated, enabling it to learn complex spatial and spectral relationships that are critical for accurate bathymetric estimation. While smaller models like MTL (15.4M) or JSH-Net (21.6M) are more computationally efficient, they exhibit inferior performance, especially when applied to medium- and low-resolution data.

To evaluate the applicability of the proposed model for real-time shallow water bathymetry retrieval applications, we assess the inference times using Sentinel-2 data over the Agia Napa area. As shown in Table 6, despite Seabed-Net having a substantially higher parameter count compared to MTL, its inference times are comparable. Specifically, Seabed-Net achieves inference times of 1.99 seconds for depth prediction and 5.17 seconds for pixel-based classification, which are only approximately 3.6% and 16.2% higher, respectively, compared to MTL's inference times. This marginal difference indicates that, while Seabed-Net requires more offline time for training due to its size, its online inference performance remains efficient and well suited for near-real-time applications.

Importantly, the specific experiments presented in Table 6 were carried out using an NVIDIA A5000 24GB GPU, a more accessible and cost-effective option compared to the A100 80GB used for training and testing. This demonstrates that high-accuracy results and near-real-time inference are achievable without requiring prohibitively expensive hardware, thus enhancing the practicality and scalability of the proposed model for broader deployment.

These results highlight the benefits of increased model capacity in challenging aquatic environments. To further improve near-real-time performance, especially in operational settings, optimizations such as efficient patch sampling, parallel processing, or cloud-based deployment can be implemented. These enhancements would allow scalable, responsive, and accurate bathymetric mapping suitable for time-sensitive coastal monitoring and management tasks.

### 5.7. Limitations and error sources

Despite Seabed-Net's strong performance, several unrelated discrepancies between predicted and reference depths contribute to residual errors, as discussed below.

One critical issue involves the inherent variability and uncertainty in the reference bathymetry itself. Residual errors in reference data can arise from multiple sources, including sensor noise, water surface dynamics, and limited underwater visibility (Agrafiotis and Demir, 2025). In Agia Napa, the standard deviation in LiDAR-derived depth measurements reaches ±0.15 m for depths up to 1.5 Secchi depth, while in the shallower and more controlled Puck Lagoon, the error was smaller at ±0.07 m (Janowski et al., 2024).

In this study, images, bathymetry, and seabed sample data were acquired with a small temporal difference of a few weeks or months (Agrafiotis et al., 2024) to minimize environmental changes and ensure visual consistency in seabed features and water conditions across sensors. This reflects real operational scenarios where multi-source data are rarely acquired on the same date. This choice enables training models that are robust to minor temporal variations, increasing their practical deployment potential. However, larger temporal differences in acquisition of remote sensing imagery and reference bathymetry may introduce further uncertainty, especially in dynamic environments. Seasonal changes in seagrass, sediment movement, or shoreline erosion can also lead to discrepancies. Although deep learning models may partially compensate for these changes through statistical learning, persistent residual errors can affect both bathymetric and pixel-based classification outputs. This is particularly relevant in nearshore zones with wave activity or wetting-drying effects.

A further, often overlooked, source of error arises from the orthorectification process used for satellite imagery. Over land, DSMs are routinely used to account for elevation changes during orthorectification. However, over water bodies, no such elevation model is typically available. As a result, standard orthorectification pipelines apply a flat-surface assumption across water areas. While this simplification enables geometric correction in the absence of bathymetric data, it introduces systematic distortions where the seafloor is sloped or irregular. This is particularly problematic in nearshore environments, where accurate pixel geolocation is essential for bathymetric modeling. The result is a mismatch between the true underwater geometry and the orthorectified image, leading to positional errors. These distortions propagate into the training and inference stages of models like Seabed-Net, which rely on precise spatial correspondence between spectral inputs and depth labels.

## 6. Conclusion

This study presented Seabed-Net, a novel multi-task deep learning framework that jointly performs bathymetry estimation and pixel-based seabed classification using remote sensing imagery. By integrating localized (via attention-based fusion) and global (via Swin Transformer) feature interactions across dual-task pathways, Seabed-Net effectively captures both geometric and semantic context, crucial for robust shallow-water mapping.

Comprehensive evaluations on the MagicBathyNet dataset across two contrasting coastal environments (Agia Napa and Puck Lagoon) demonstrated that Seabed-Net consistently outperforms both single-task and multi-task models. For example, the model achieved up to a 38% reduction in RMSE compared to single-task bathymetry networks under Sentinel-2 imagery (from 1.068 m to 0.660 m), and improved seabed classification accuracy by up to 16% in the same low-resolution settings. In Puck Lagoon, Seabed-Net reached an overall classification accuracy of 98.00%, outperforming all baselines across all sensors. Additionally, Seabed-Net substantially surpassed traditional empirical bathymetry models, such as Multiband Linear (Lyzenga, 1978) and Log Band Ratio (Stumpf et al., 2003), achieving RMSE reductions of 40-60% for aerial data, 60-70% for SPOT 6, and 65-75% for Sentinel-2 imagery, with the largest performance gaps observed in the complex Agia Napa region. In comparison to traditional machine learning regressors like Random Forest and Support Vector Regression, Seabed-Net also delivered significantly lower RMSE, MAE, and standard deviation values across all datasets. For instance, in Agia Napa, Seabed-Net reduced RMSE by 24-42% over Random Forest and 32-49% over SVR, while in Puck Lagoon, it maintained lower errors with gains ranging from 14-39%. Notably, Seabed-Net achieved a 0.43 m RMSE on Sentinel-2 imagery in Puck Lagoon, outperforming Random Forest and SVR (0.642 m and 0.637 m, respectively). Performance gains are particularly evident in structurally diverse or low-contrast regions, where shared semantic and geometric cues help mitigate common misinterpretations of depth and substrate. In addition, CATZOC-based evaluations indicate that Seabed-Net's depth predictions meet high-confidence thresholds (A1 or A2) in more than 90% of cases across all sensors.

These consistent improvements underline the superiority of Seabed-Net's deep multitask architecture in capturing complex spatial-spectral relationships, modeling non-linearities, and generalizing effectively across diverse environments. Crucially, the model maintained strong performance across regions of varying size and data density, showing comparable effectiveness in the smaller Agia Napa site as in the larger Puck Lagoon. This robustness underscores the model's ability to generalize well, regardless of dataset scale, bathymetric complexity, or geographic variability.

To the best of our knowledge, Seabed-Net is the first multi-task learning framework in which both bathymetry estimation and pixel-wise seabed classification are treated as final supervised tasks, rather than auxiliary or intermediate ones. This joint optimization strategy, combined with dynamic task uncertainty weighting, enables synergistic learning that improves performance in complex optical environments. While our model requires paired in-situ bathymetry and habitat/ substrate labels during training, this supervision is only needed for training; once trained, the model can generate both bathymetric and benthic classification maps from remote sensing imagery alone, without the need for additional in-situ data at inference time. This setup, while more demanding during training, enables fully automated, end-to-end deployment in data-scarce regions. This contrasts with most existing methods, which address either bathymetry or seafloor classification in isolation and often lack the capacity to generalize jointly across both.

Although the current architecture is relatively large, the demonstrated inference efficiency, achieved even on accessible hardware, shows that Seabed-Net is already viable for near-real-time applications. Future work will focus on developing parameter-efficient variants, adaptive scaling strategies, and domain adaptation techniques to further improve deployment flexibility and reduce training overhead. Our model, like all supervised models, implicitly learns the complex relationships between light, water, and the seabed directly from the data through joint training. This can offer greater flexibility and scalability in operational, multi-sensor scenarios, where explicitly modeling physical processes may be impractical. However, we also plan to explore the integration of physics-informed components, such as radiative transfer constraints or light attenuation models, to explicitly encode these interactions and potentially improve generalization, interpretability, and robustness in optically complex environments. Overall, Seabed-Net lays the groundwork for integrated, scalable remote sensing tools for shallow-water environments, where both structural and semantic accuracy are critical.

## Acknowledgment

This work is part of MagicBathy: Multimodal multi-tAsk learninG for MultIsCale BATHYmetric mapping in shallow waters project funded by the European Union's HORIZON Europe research and innovation programme under Marie Skłodowska-Curie Actions with Grant Agreement No. 101063294. Views and opinions expressed are, however, those of the authors only and do not necessarily reflect those of the European Union and the granting authority. Neither the European Union nor the granting authority can be held responsible for them.

## CRediT authorship contribution statement

**Panagiotis Agrafiotis**: Conceptualization, Data curation, Formal analysis, Funding acquisition, Methodology, Software, Validation, Visualization, Writing - original draft, Writing - review & editing. **Begüm Demir**: Supervision, Writing - review & editing.

## Appendix A. Supplementary Material

This appendix contains supplementary material supporting the main text. Specifically, it presents the overall results of bathymetry and pixel-based classification using SeabedNet at all the test sites.